\documentclass[runningheads]{llncs}
\usepackage{graphicx}
\usepackage{tikz}
\usepackage{comment}
\usepackage{amsmath,amssymb} 
\usepackage{color}

\usepackage[font=small]{subfig}
\usepackage{comment}
\usepackage{multirow}
\usepackage{booktabs}
\usepackage{xcolor, colortbl}
\usepackage{tabularx}
\usepackage[super]{nth}
\newcolumntype{Y}{>{\centering\arraybackslash}X}
\usepackage{enumitem}
\usepackage{algorithm}
\usepackage{algorithmic}
\usepackage{color,soul}

\begin{document}
\pagestyle{headings}
\mainmatter
\def\ECCVSubNumber{982}  

\title{Unsupervised Night Image Enhancement: \\
	When Layer Decomposition Meets \\ Light-Effects Suppression} 

\titlerunning{Unsupervised Night Image Enhance.: Layer Decomp. Meets L.E. Supp.}
%
\author{Yeying Jin\inst{1}\orcidID{0000-0001-7818-9534} \and
Wenhan Yang\inst{2}\orcidID{0000-0002-1692-0069} \and
Robby T. Tan\inst{1,3}\orcidID{0000-0001-7532-6919}}
\authorrunning{Y. Jin, W. Yang et al.}
%
\institute{National University of Singapore \and Nanyang Technological University \and Yale-NUS College\\
\email{jinyeying@u.nus.edu, wenhan.yang@ntu.edu.sg, robby.tan@\{nus,yale-nus\}.edu.sg}}
\maketitle

\begin{abstract}
	Night images suffer not only from low light, but also from uneven distributions of light. 
	Most existing night visibility enhancement methods focus mainly on enhancing low-light regions. This inevitably leads to over enhancement and saturation in bright regions, such as those regions affected by light effects (glare, floodlight, etc).
	To address this problem, we need to suppress the light effects in bright regions while, at the same time, boosting the intensity of dark regions.
	With this idea in mind, we introduce an unsupervised method that integrates a layer decomposition network and a light-effects suppression network.
	Given a single night image as input, our decomposition network learns to decompose shading, reflectance and light-effects layers, guided by unsupervised layer-specific prior losses.
	Our light-effects suppression network further suppresses the light effects and, at the same time, enhances the illumination in dark regions.
	This light-effects suppression network exploits the estimated light-effects layer as the guidance to focus on the light-effects regions.
	To recover the background details and reduce hallucination/artefacts, we propose structure and high-frequency consistency losses.
	Our quantitative and qualitative evaluations on real images show that our method outperforms state-of-the-art methods in suppressing night light effects and boosting the intensity of dark regions.
	\footnote{Our data and code is available at: \url{https://github.com/jinyeying/night-enhancement}}
	\keywords{Night image enhancement, low-light image, light-effects suppression}
\end{abstract}

\section{Introduction}
\label{sec:intr}
\begin{figure}[t]
	\captionsetup[subfloat]{labelformat=empty}
	\captionsetup[subfloat]{farskip=2pt}
	\subfloat[Input]{\includegraphics[width=0.245\textwidth]{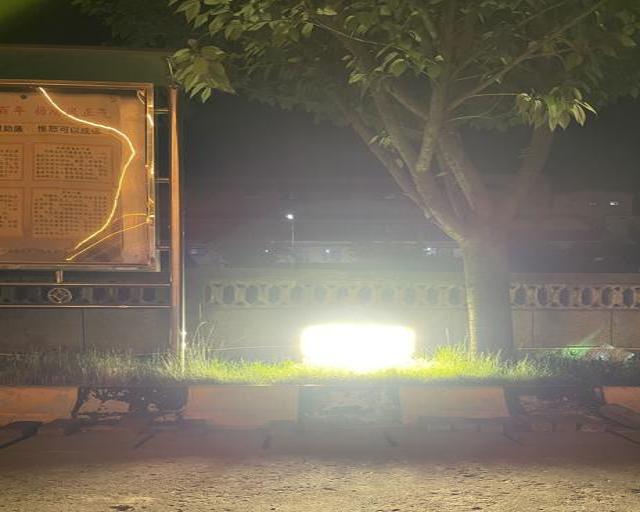}}\hfill
	\subfloat[Our Method]{\includegraphics[width=0.245\textwidth]{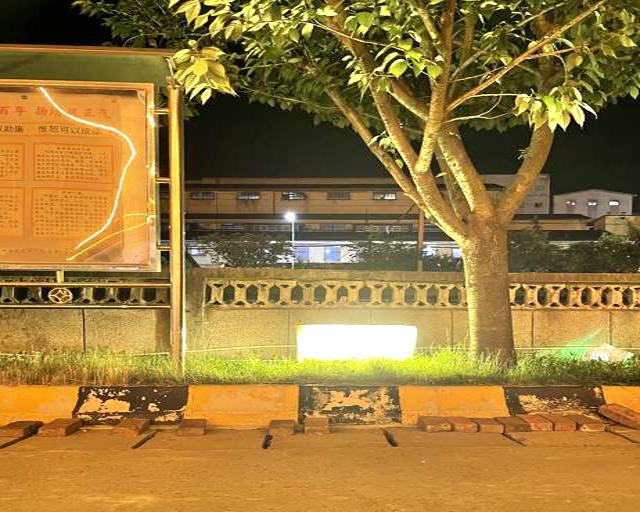}}\hfill
	\subfloat[Sharma~\cite{sharma2021nighttime}]{\includegraphics[width=0.245\textwidth]{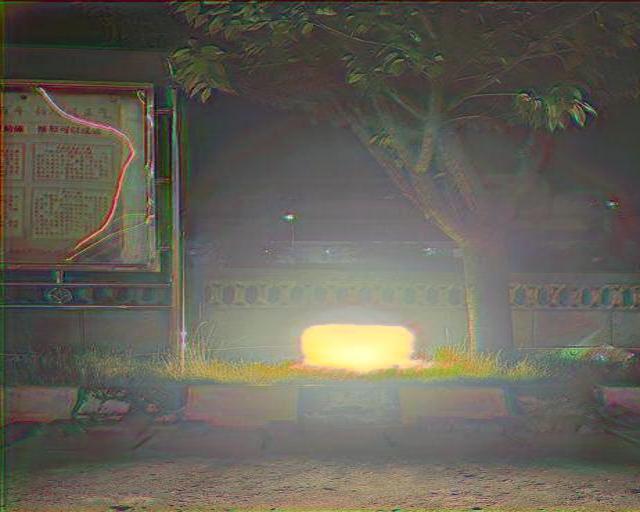}}\hfill
	\subfloat[EnlightenGAN~\cite{jiang2021enlightengan}]{\includegraphics[width=0.245\textwidth]{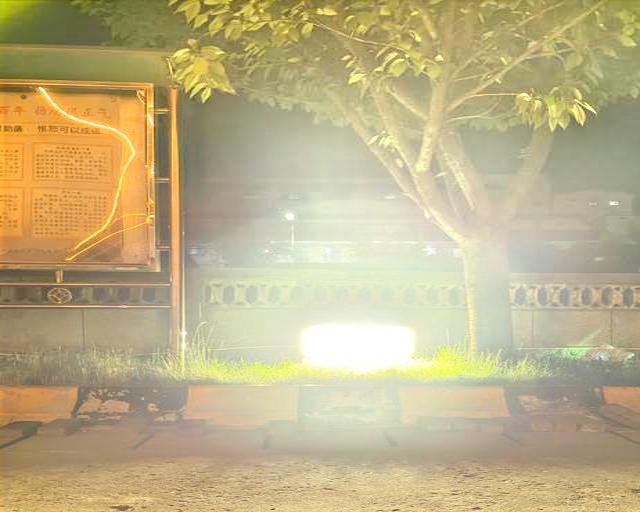}}\hfill
	\caption{An existing night light-effects suppression method~\cite{sharma2021nighttime} suffers from hallucination/artefacts and generates improper light effects, while an image enhancement method~\cite{jiang2021enlightengan} is not designed to handle night light effects and incorrectly intensifies it. In contrast, our method jointly suppresses light effects and enhances dark regions.}
	\label{figure_intro}
\end{figure}

Night images can contain uneven light distributions, as shown in Fig.~\ref{figure_intro}, where some regions are dark and some are significantly brighter, due to the presence of light effects\footnote{Following~\cite{sharma2021nighttime}, light effects in this paper refer to glare, floodlight, etc.}.
Most existing nighttime visibility enhancement methods focus mainly on boosting the intensity of low-light regions, e.g.,~\cite{guo2016lime,chen2018retinex,wang2019underexposed,guo2020zero,jiang2021enlightengan}.
Hence, when these methods are applied to night images that contain light effects, they inevitably amplify the light effects, and impair the visibility of the images even further. Unlike these methods, our goal in this paper is to suppress the light effects while, at the same time, boosting the intensity of dark regions.

Fully-supervised learning methods could be a possible solution to achieving our goal. However, these methods would require a diverse and large collection of paired night images taken with and without light effects, which is intractable to obtain. 
Another possible solution would be the use of synthetic night images with rendered light effects. However, the effectiveness of methods trained on synthetic night data depends on the quality of the light-effects rendering model. To our knowledge, rendering physically correct night light effects with various background scenes and lighting conditions is still challenging~\cite{wu2021train}.

In this paper, we introduce an unsupervised learning approach that integrates a decomposition network and a light-effects suppression network in a single unified framework.
Our decomposition network is derived from an image-layer model and guided by our layer-specific prior losses to decompose the input image into shading, reflectance and light-effects layers (Fig.~\ref{figure_decomposition} shows the examples of these three layers).
Subsequently, our light-effects suppression network, which is trained on unpaired images with and without light effects, provides additional unsupervised constraints. 
This network not only strengthens the light effects decomposition but also enhances the intensity in dark regions.
The two networks, the decomposition and light-effect suppression networks, are connected.

To recover the background details behind light-effects regions, we introduce structure and high-frequency (HF) features consistency losses.
We employ the structure consistency based on the VGG network and utilize the guided filter to obtain HF features.
The structure and HF-features consistency losses can also reduce hallucination.
In summary, our main contributions are as follows:
\begin{itemize}
	\item 
	To enhance the visibility of night images that suffer from low light and light effects simultaneously, we introduce a network architecture that integrates layer decomposition and light-effects suppression in one unified framework.
	\item
	To distinguish light effects from background regions, particularly when the color of the light effects is white or achromatic, we propose utilizing the estimated light-effects layer as guidance for our unsupervised light-effects suppression network.
	\item 	
	To restore the background details, we introduce novel unsupervised losses based on the structure and HF-features consistency.
	Our perceptual structure information and HF texture information are less affected by light effects. Thus, they can be employed to preserve background details, and, importantly, to suppress unwanted artefacts.
\end{itemize}
Our experiments and evaluations show that our method is effective in suppressing light-effects regions and enhancing dark regions, outperforming state-of-the-art methods both quantitatively and qualitatively.

\section{Related Work}
\label{sec:related_work}
Sharma and Tan~\cite{sharma2021nighttime} introduce a method based on camera response function (CRF) estimation and HDR imaging to suppress light effects. The method is the first method that can suppress light effects and improve the dynamic range for night images.
However, it suffers from artefacts and missing details as shown in Fig.~\ref{figure_intro}, particularly for white (or achromatic) lights.

In the field of night image dehazing, a few methods have been proposed to suppress glow due to haze/fog particles.  
Li et al.~\cite{li2015nighttime} address glow removal on foggy nights using layer separation.
Zhang et al.~\cite{zhang2017fast} use maximum reflectance prior for haze and glow removal.
Ancuti et al.~\cite{ancuti2016night,ancuti2020day} use a fusion process and the Laplace operator to deglow and dehaze.
Yan et al.~\cite{yan2020nighttime,yan2021self} propose a semi-supervised method~\cite{yan2020optical} employing a grayscale guided network.
However, all these methods are designed for glow suppression in haze or foggy night, and not for removing light effects in clear night images. Moreover, unlike our method, they are also not designed for enhancing dark regions.

A number of methods have been developed to boost the brightness of low-light images without considering the presence of night light effects. A few methods are based on histogram equalization~\cite{pizer1987adaptive}, inversion and dehazing~\cite{dong2011fast}, the retinex model (e.g.~\cite{fu2016weighted,li2018structure}), while more recent methods are based on deep networks~\cite{li2021low}.
Most deep-learning-based methods (e.g.~\cite{afifi2021learning,chen2018retinex,wang2019underexposed}) adopt supervised learning to train their model and thus require a large number of pairs of low/normal-light images.
A few unsupervised methods (e.g.~\cite{jiang2021enlightengan}) rely on adversarial training using unpaired low/normal-light images.
Semi-supervised methods (e.g.~\cite{yang2020fidelity,yang2021band}) recompose coarse-to-fine representations towards perceptually pleasing images with the help of unpaired high-quality images.
Recently, zero-shot learning methods (e.g.~\cite{li2021learning,guo2020zero}) have been proposed for low-light enhancement. 
Most of these night image enhancement methods, however, are not designed to suppress night light effects and enhance low light regions simultaneously; therein lies the main difference with our work.

\section{Proposed Method}
\label{sec:method}

\begin{figure}[t!]
	\captionsetup[subfloat]{farskip=1pt}
	\centering
	\subfloat{\includegraphics[width=12cm, height=4cm]{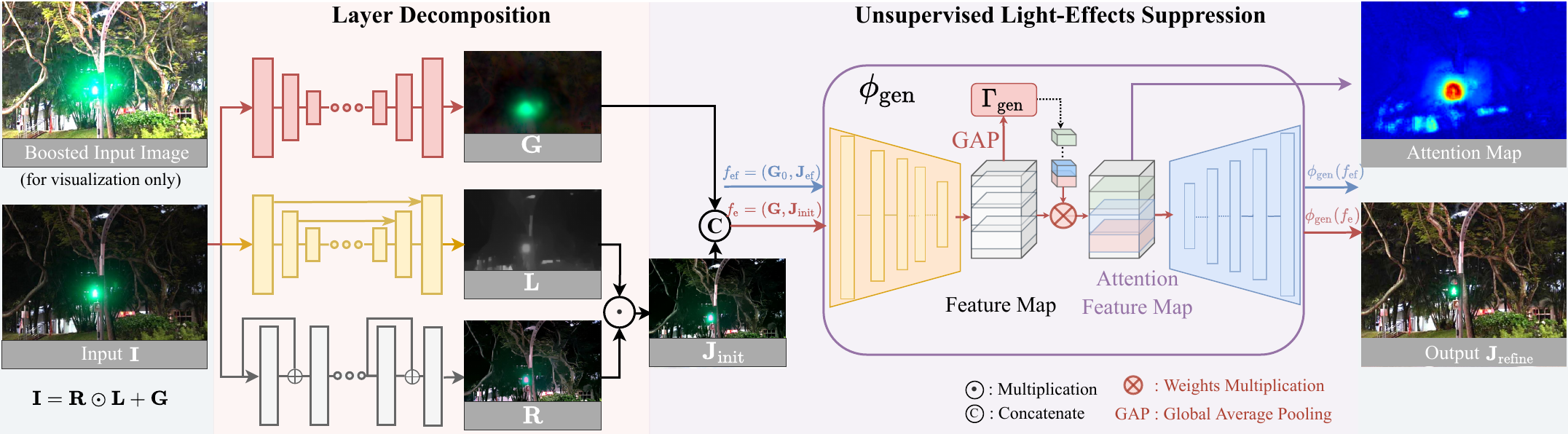}}
	\caption{The overall architecture of our proposed method.
		We integrate decomposition and light-effects suppression networks in one unified unsupervised framework.
		Given the input night image, we suppress light effects through the layer decomposition network, in which light-effects, shading, and reflectance layers are obtained (see Fig.~\ref{figure_decomposition}). 
		The light-effects suppression is guided by the decomposed light-effects layer $\mathbf{G}$ and based on unpaired learning (see Fig.~\ref{figure_attention}) to further suppress light effects and boost dark regions.}
	\label{figure_model}
\end{figure}

To suppress light effects and, at the same time, boost the intensity of dark regions, we propose an unsupervised framework by integrating a decomposition network and a light-effects suppression network.
Our decomposition network is based on an image-layer model and produces three separate layers: shading, reflectance, and light-effects layers.
We input these layers into our light-effects suppression network to obtain our final output, where light effects are suppressed and dark regions are boosted. 
This network learns from unpaired data and is guided by our estimated light-effects layer.

\subsection{Model-Based Layer Decomposition Network}
Our decomposition is based on the following image-layer model:
\begin{equation}
\mathbf{I} = \mathbf{R} \odot \mathbf{L} + \mathbf{G},
\label{eq:retinex_our}
\end{equation}where $\mathbf{I}$ represents the input night image, $\mathbf{G}$ represents the light-effects layer, $\mathbf{R}$ and $\mathbf{L}$ are the reflectance and shading layers, respectively. The notation $\odot$ represents element-wise multiplication. 
In this equation, we assume a linear gamma function. However, we do not use this equation explicitly in our method. Instead, we use it only to guide the design of our network in Fig.~\ref{figure_model} (i.e., the layer decomposition network). When non-linear images with non-linear gamma functions are used in training, the background scenes are approximations of the physically correct values.
Our decomposition goal is to obtain a background scene that is free from light effects, i.e., we want to estimate the background scene, $\mathbf{J}_\text{init}=\mathbf{R}\odot\mathbf{L}$.
Hence, even when non-linear images are used in training, applications that are less concerned about physically correct intensity values but suffer from light effects can benefit from our method. 
Our model differs from the widely used intrinsic model~\cite{gehler11nips,bell2014intrinsic}, as the latter does not incorporate the light-effects layer.

Fig.~\ref{figure_model} shows our pipeline. The decomposition network is based on our image-layer model in Eq.~(\ref{eq:retinex_our}). 
Given the input image ($\mathbf{I}$), we first perform image decomposition. We use three separate networks and our novel unsupervised losses to obtain the light effects ($\mathbf{G}$), shading ($\mathbf{L}$), and reflectance ($\mathbf{R}$) layers.

\begin{figure}[t]
	\centering
	\captionsetup[subfloat]{farskip=1pt}
	\setcounter{subfigure}{0}
	\subfloat[Input $\mathbf{I}$]{\includegraphics[width = 0.245\columnwidth]{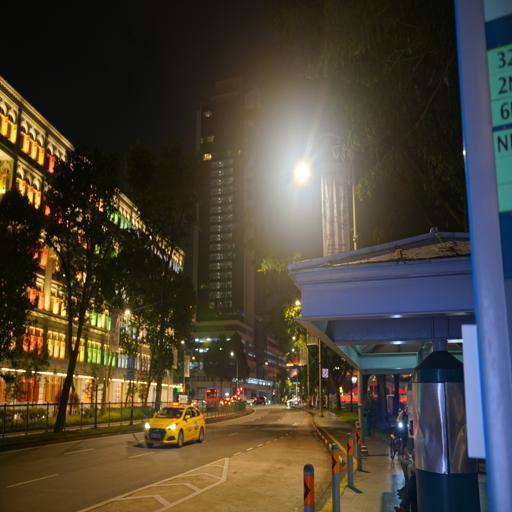}}\hfill
	\subfloat[$\mathbf{G}$\label{fig_G}]{\includegraphics[width = 0.245\columnwidth]{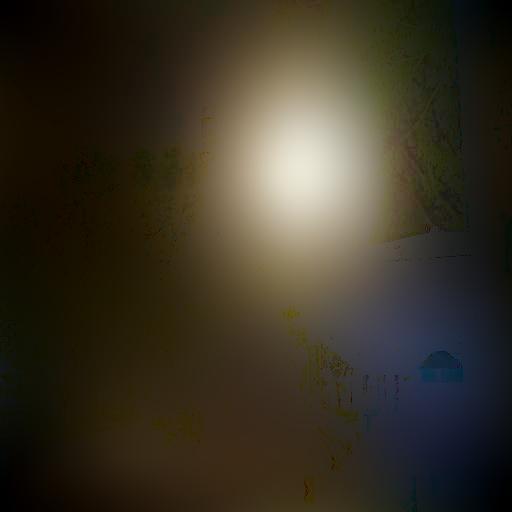}}\hfill
	\subfloat[$\mathbf{L}$]{\includegraphics[width = 0.245\columnwidth]{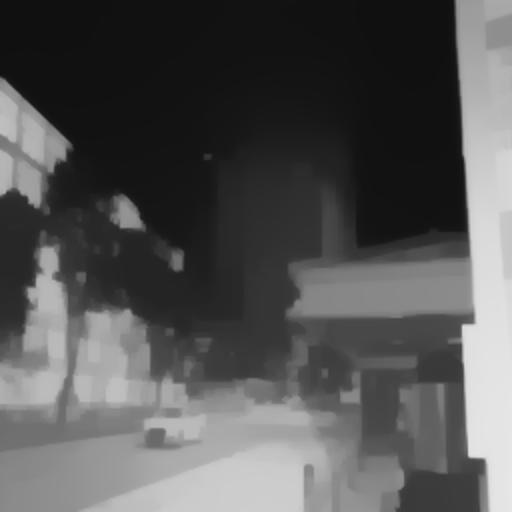}}\hfill
	\subfloat[$\mathbf{R}$]{\includegraphics[width = 0.245\columnwidth]{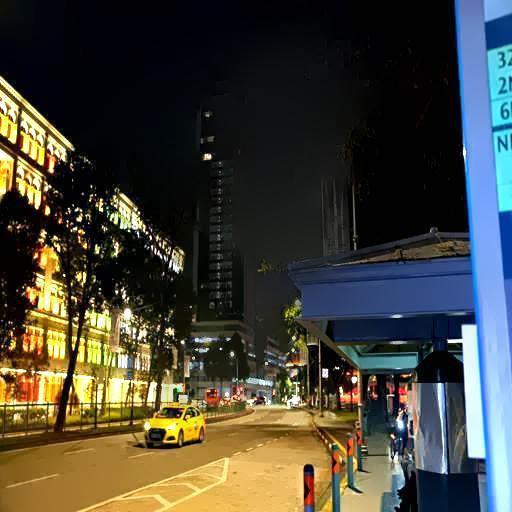}}\hfill\\
	\caption{Results of our model-based layer decomposition. (a) Input. (b) Light-effects layer $\mathbf{G}$. (c) Shading layer $\mathbf{L}$ and (d) Reflectance layer $\mathbf{R}$.}
	\label{figure_decomposition}
\end{figure}

\subsubsection{Learning Light Effects, Shading and Reflectance Layers}
To obtain the light effects ($\mathbf{G}$), shading ($\mathbf{L}$), and reflectance ($\mathbf{R}$) layers, we use three networks respectively: Light-Effects-Net ($\phi_\text{G}$), Shading-Net ($\phi_\text{L}$) and Reflectance-Net ($\phi_\text{R}$), where 
$\mathbf{G}=\phi_\text{G}\left(\mathbf{I}\right)$, $\mathbf{L}=\phi_\text{L}\left(\mathbf{I}\right)$, and $\mathbf{R}=\phi_\text{R}\left(\mathbf{I}\right)$.
The three networks are trained using unsupervised losses, which will be discussed in the subsequent paragraphs. 
Fig.~\ref{figure_decomposition} shows examples of these three layers.

\subsubsection{Light-Effects and Shading Initialization}
To resolve the decomposition ambiguity problem, it is important to provide proper initial estimates of the layers. 
For the shading layer, we employ a shading map $\mathbf{L}_{\text{i}}$ obtained by taking the maximum value of the three color channels, for each pixel~\cite{guo2016lime}. 
For the light-effects layer, we use a light-effects map $\mathbf{G}_{\text{i}}$, computed using 
the relative smoothness technique~\cite{li2014single}. This is extracted using the second-order Laplacian filter from the input image, since light effects are smooth variations.
We define the loss function for the initialization step as:
\begin{equation}
\setlength{\abovedisplayskip}{1pt}
\setlength{\belowdisplayskip}{1pt}
\mathcal{L}_\text{init} = |\mathbf{G} - \mathbf{G}_{\text{i}} |_{1} + |\mathbf{L} - \mathbf{L}_{\text{i}}|_{1}. 
\label{eq:loss_g_l}
\end{equation}

\subsubsection{Gradient Exclusion Loss}
The gradients of the light effects layer have a short tail distribution, similar to that of 'glow'~\cite{li2015nighttime}. In contrast, the gradients of the background image have a long tail distribution~\cite{li2014single}.
Hence, we employ a gradient exclusion loss to recover the uncorrelated layers $\mathbf{\{G,\mathbf{J}_\text{init}}\}$, where the goal is to separate the two layers as far as possible in the gradient space.
The definition of the loss follows~\cite{gandelsman2019double,zhang2018single}: 
\begin{equation}
\setlength{\abovedisplayskip}{1pt}
\setlength{\belowdisplayskip}{1pt}
\mathcal{L}_\text{excl} = \sum_{n=1}^{3}\big\lVert \mathbf{\tanh}(\lambda_{\mathbf{G}^{\downarrow\!n}}\lvert\nabla\mathbf{G}^{\downarrow\!n}\rvert)\circ \mathbf{\tanh}(\lambda_{\mathbf{J}^{\downarrow\!n}_\text{init}}\lvert\nabla\mathbf{J}^{\downarrow\!n}_\text{init}\rvert)\big\rVert_{F},
\end{equation}where $\lVert\cdot\rVert_{F}$ is the Frobenius norm, $\mathbf{G}^{\downarrow\!n}$ and $\mathbf{J}^{\downarrow\!n}_\text{init}$ represent $\mathbf{G}$ and $\mathbf{J}_\text{init}$ downsampled using the bilinear interpolation, and the parameters $\lambda_{\mathbf{G}^{\downarrow\!n}}\!$ and $\lambda_{\mathbf{J}^{\downarrow\!n}_\text{init}}\!$ are normalization factors. 

\begin{figure}[t!]
	\captionsetup[subfloat]{farskip=1pt}
	\centering
	{\includegraphics[width=1\textwidth]{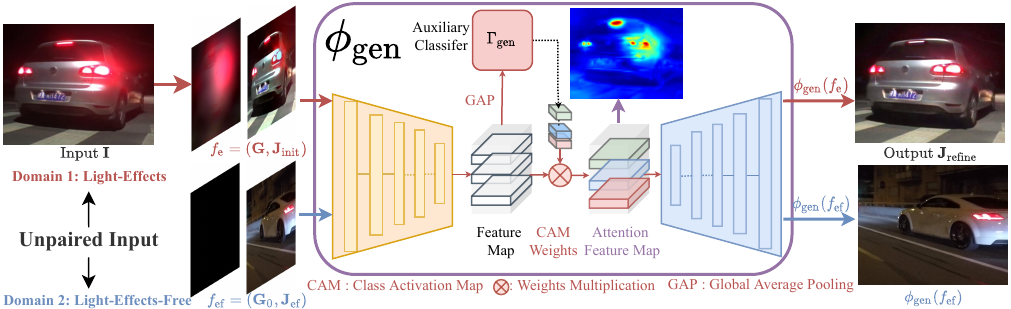}}
	\caption{Overview of our unsupervised light-effects suppression network.
		The network comprises a generator $\phi_\text{gen}$ and a classifier $\Gamma_\text{gen}$.
		The encoder block of our generator extracts feature maps from the input image layers. Our classifier $\Gamma_\text{gen}$ is trained to learn the weights~\cite{zhou2016learning} of the feature maps. 
		$\Gamma_\text{gen}$ performs domain classification based on two domains, i.e., the light-effects domain $f_\text{e}=\left(\mathbf{G}, \mathbf{J}_\text{init}\right)$ and the unpaired light-effects-free domain $f_\text{ef}=\left(\mathbf{G}_\text{0},\mathbf{J}_\text{ef}\right)$.
		Averaging the weighted feature maps generates an attention map that shows the network is focusing on the light-effects regions.
		As a result, the light effects are significantly suppressed in our output $\mathbf{J}_\text{refine}$.}
	\label{figure_attention}
\end{figure}

\subsubsection{Color Constancy Loss}
To minimize any color shift in our decomposition output, inspired by the Gray World  assumption~\cite{buchsbaum1980spatial,guo2020zero,sharma2021nighttime}, we use a color-constancy prior, which encourages the range of the intensity values of the three color channels in the background image $\mathbf{J}_\text{init}$ to be balanced:
\begin{equation}
\label{eq_testtime_grayloss}
\setlength{\abovedisplayskip}{1pt}
\setlength{\belowdisplayskip}{1pt}
\mathcal{L}_\text{cc} = \sum_{(c1,c2)} \big(| \mathbf{J}_\text{init}^{c1}  - \mathbf{J}_\text{init}^{c2}|_1\big), 
\end{equation}where $(\text{c1},\text{c2})\in\{(r,g),(r,b),(g,b)\}$ denotes a combination of two color channels.

\subsubsection{Reconstruction Loss}
For our decomposition task, recombining the estimated layers should give us back the original input image.
Hence, we define our reconstruction loss as: 
\begin{equation}
\setlength{\abovedisplayskip}{1pt}
\setlength{\belowdisplayskip}{1pt}
\mathcal{L}_\text{recon} = |\mathbf{I}  - (\mathbf{R} \odot \mathbf{L} + \mathbf{G}\big)|_1.
\label{eq_testtime_reconloss}
\end{equation}
We multiply each unsupervised loss with its respective weight, where we set $\lambda_\text{init}$, $\lambda_\text{excl}$ all set to 1 since they are in the same scale.
We empirically set $\lambda_\text{recon}=0.1$ and employ the weight $\lambda_\text{cc}=0.5$ from~\cite{guo2020zero} to balance the decomposition process.

\subsection{Light-Effects Suppression Network}
To better suppress light effects, we integrate our decomposition network with an unpaired light-effects suppression network. 
We design this network to suppress light effects by using the guidance of our estimated light-effects layer, enforcing the network to focus on light-effects regions.  
As shown in Fig.~\ref{figure_model}, our network comprises a generator $\phi_\text{gen}$ and a classifier $\Gamma_\text{gen}$.
It refines the initially estimated background scene ($\mathbf{J}_\text{init}$), and generates the final light-effects-free output ($\mathbf{J}_\text{refine}$). 
The details are as follows.

\subsubsection{Light-Effects Layer Guidance}
We employ the estimated light-effects layer $\mathbf{G}$ to guide our training process, as shown in Fig.~\ref{figure_attention}.
The light-effects layer is taken as part of the input of our encoder-decoder network, and is modulated with the feature maps of the network at different scales.
Specifically, we concatenate $\mathbf{J}_\text{init}$ with the light-effects layer $\mathbf{G}$, and then we input them to our network $\phi_\text{gen}$.

By resizing the light-effects layer, $\mathbf{G}$, to fit the size of each feature map, and multiplying it with all the intermediate feature maps, our light-effects layer can guide our network to focus more on light-effects regions.
Fig.~\ref{fig_G} and Fig.~\ref{fig_ab_layer} show some results of our light-effects layers, demonstrating that our method can successfully separate white and multi-color light effects.

\begin{figure}[t!]
	\captionsetup[subfloat]{farskip=1pt}
	\centering
	{\includegraphics[width=1\textwidth]{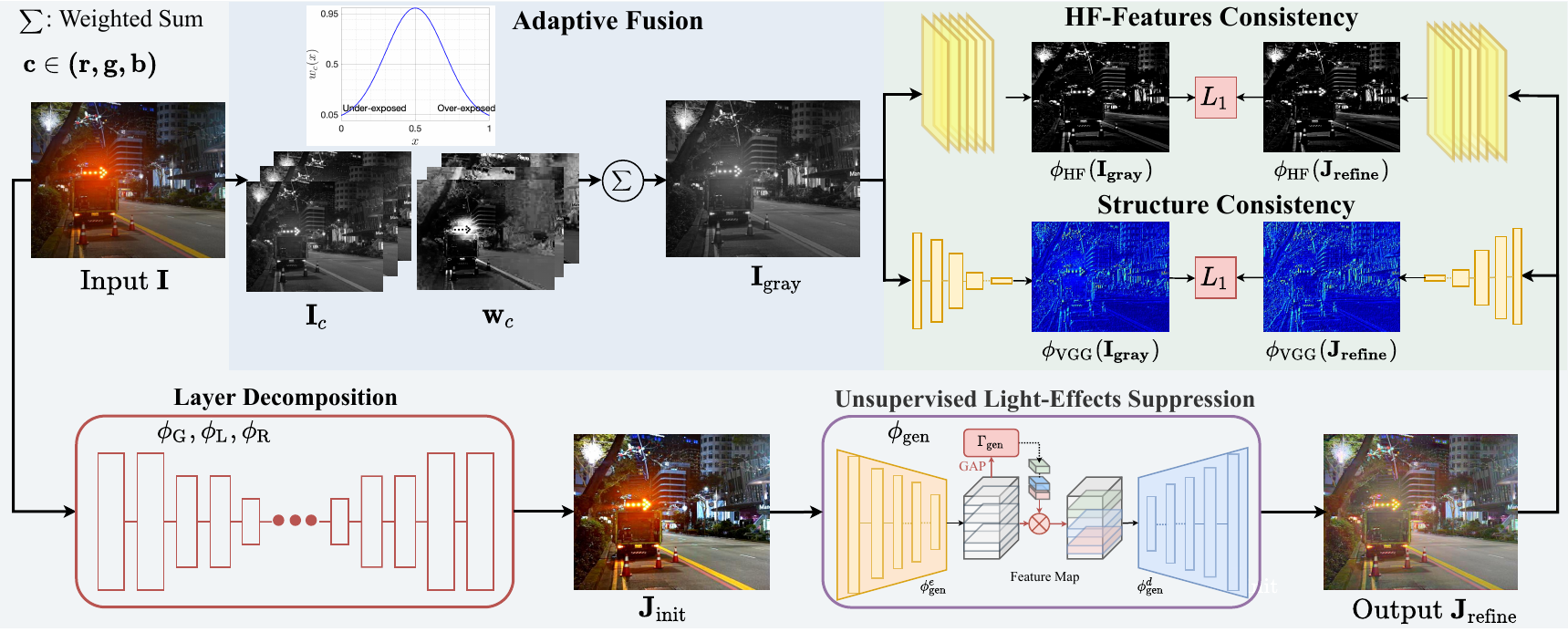}}
	\caption{Overview of our structure and HF-features consistency losses.
		We first use our adaptive fusion scheme to obtain a fused grayscale image $\mathbf{I}_\text{gray}$. 
		Then, from $\mathbf{I}_\text{gray}$, we compute VGG features $\phi_\text{VGG}(\mathbf{I}_\text{gray})$ that are less affected by light effects, and HF-features $\phi_\text{HF}(\mathbf{I}_\text{gray})$ that are more robust to light effects and contain background details.}
	\label{figure_features}
\end{figure}

\subsubsection{Light-Effects Suppression}
Besides the light-effects layer, our suppression network is also guided by an attention mechanism~\cite{jiang2021enlightengan,kim2019u,jin2021dc}.
The basic idea is that, we input the light-effects and light-effects-free unpaired images into our encoder-decoder network. We then, use a domain classifier to judge whether the encoded features come from a certain domain, i.e., to judge whether the input is light-effects or light-effects-free.
Using this domain classification, the activated feature regions can form an attention map~\cite{zhou2016learning} that is useful when guiding our network in suppressing light effects.

More specifically, as shown in Fig.~\ref{figure_attention}, our network $\phi_\text{gen}$ contains an auxiliary classifier $\Gamma_\text{gen}$. One of the inputs of the network is the concatenation of $\mathbf{J}_\text{init}$ and $\mathbf{G}$.    
Another input is a light-effects-free reference image, $\mathbf{J}_\text{ef}$, concatenated with a dummy all zero map $\mathbf{G}_\text{0}$, which of course has no light effects.
Our classifier, $\Gamma_\text{gen}$, then performs domain classification based on the encoded features from $f_\text{e}=\left(\mathbf{G}, \mathbf{J}_\text{init}\right)$ or $f_\text{ef}=\left(\mathbf{G}_\text{0},\mathbf{J}_\text{ef}\right)$.
To train the auxiliary classifier $\Gamma_\text{gen}$, we use the following attention loss:
\begin{equation}
\setlength{\abovedisplayskip}{1pt}
\setlength{\belowdisplayskip}{1pt}
\mathcal{L}_{\text{atten}} =
-\big(\mathbb{E}\big[\log(\Gamma_\text{gen}(f_\text{e}))\big]+\mathbb{E}\big[\log(1-\Gamma_\text{gen}(f_\text{ef})))\big])\big..
\label{eq:loss_cam_g}
\end{equation}

\subsubsection{Structure and HF-Features Consistency Losses}
To address hallucination/artefacts~\cite{sharma2020nighttime}, and also to preserve background details, we employ two constraints: structure consistency, based on features obtained from the VGG network~\cite{Johnson16}; and HF-features consistency, based on the HF features obtained from the guided filter~\cite{wu2018fast}.

\begin{figure}[t!]
	\captionsetup[subfloat]{labelformat=empty}
	\captionsetup[subfloat]{farskip=1pt}
	\centering
	\subfloat[Input $\mathbf{I}$]{\includegraphics[width = 0.196\columnwidth]{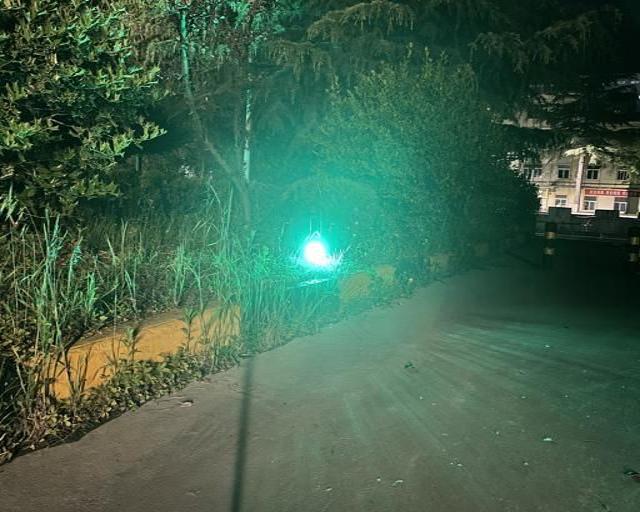}}\hfill
	\subfloat[$\mathbf{I}_\text{gray}$]{\includegraphics[width = 0.196\columnwidth]{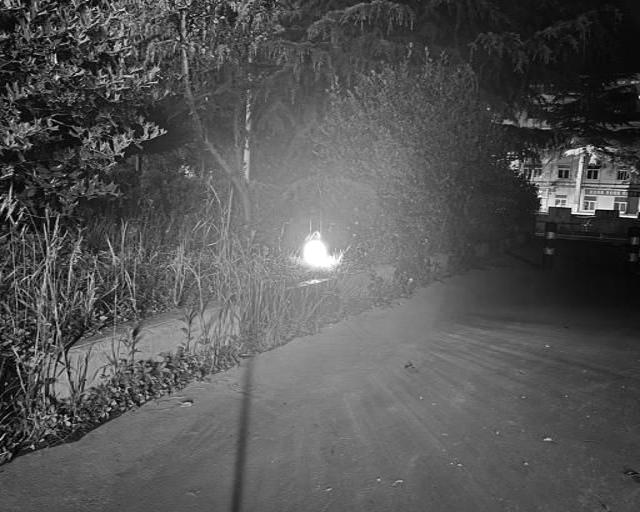}}\hfill
	\subfloat[$\phi_\text{VGG}(\mathbf{I}_\text{gray})$]{\includegraphics[width = 0.196\columnwidth]{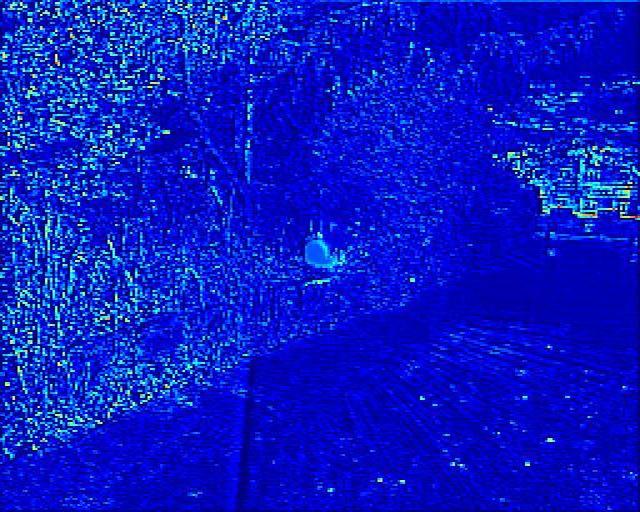}}\hfill
	\subfloat[$\phi_\text{HF}(\mathbf{I}_\text{gray})$]{\includegraphics[width = 0.196\columnwidth]{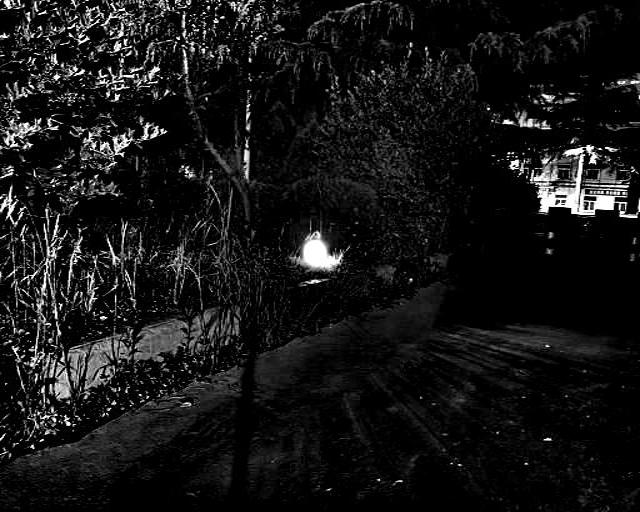}}\hfill
	\subfloat[$\mathbf{J}_\text{refine}$]{\includegraphics[width = 0.196\columnwidth]{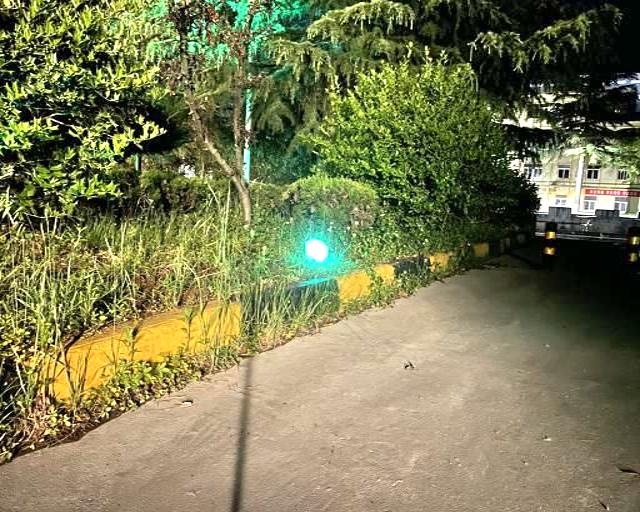}}\hfill
	\caption{Examples of feature map from VGG for $\mathbf{I}_\text{gray}$, and a HF feature map for $\mathbf{I}_\text{gray}$. As one can observe, these features are less affected by light effects.}
	\label{figure_vgg_features}
\end{figure}

As shown in Fig.~\ref{figure_features}, to obtain the structure information and HF-features that are more robust to light effects, we adaptively fuse the RGB color channels of the input night image by applying: $I_\text{gray}(\text{x}) = \sum_{c}\frac{1}{3}(w_c(\text{x})I_c(\text{x}))$ where $c\in{(r, g, b)}$ is a color channel, $\text{x}$ is a pixel location, and the input image $\mathbf{I} = \{I_r, I_g, I_b\}$.
The weight map for each color channel of the night image $I_c(\text{x})$ is computed by $w_c(\text{x})= \exp\left(\frac{-(I_c(\text{x})-0.5)^{2}}{2\sigma^{2}}\right)$. Note that the range of $I_c(\text{x})$ is [0,1], thus 0.5 is the median of the intensity range. 
Our weight has a low value if a pixel in a color channel is either low (under-exposed) or high (e.g., a light-effects pixel).
We define $\sigma=0.2$, which measures how well-exposed a pixel is.
This makes the resulting grayscale image $I_\text{gray}$ less affected by light effects, as can be observed in Fig.~\ref{figure_features} and Fig.~\ref{figure_vgg_features}.

Having obtained $I_\text{gray}$, we define our loss as follows:
\begin{equation}
\mathcal{L}_\text{gray-feat} = 
\big |\phi_\text{HF}(\mathbf{J_\text{refine}})  - \phi_\text{HF}(\mathbf{I_\text{gray}}) \big|_1 \nonumber
+
\\
\big| \phi^l_\text{VGG}(\mathbf{J_\text{refine}})  - \phi^l_\text{VGG}(\mathbf{I}_\text{gray}) \big|_1, 
\label{eq_gray}
\end{equation}where $\mathbf{I_\text{gray}} = \{I_\text{gray}, I_\text{gray}, I_\text{gray}\}$. 
$\phi^l_\text{VGG}(.)$ represents the feature maps extracted from the $l^\text{th}$ layer of the VGG16 network (we set $l=15$ in our experiments).
$\phi_\text{HF}(.)$ represent the high-frequency feature maps obtained from the guided filter. 
We concatenate these HF layers to get $\phi_\text{HF}(\mathbf{I}_\text{gray})$. 
We use these features to better preserve the HF information in the generated refined background image $\mathbf{J}_\text{refine}$.
Fig.~\ref{figure_features} shows our adaptive fusion scheme to obtain $\mathbf{I}_\text{gray}$ from which we compute HF-features and VGG-features.  
Fig.~\ref{figure_vgg_features} shows that with our loss in place, the VGG and HF features of $\mathbf{I}_\text{gray}$ preserve the structural information.

\subsubsection{Adversarial and Identity Losses}
Our adversarial loss for the generator and discriminator $\phi_\text{dis}$ uses its standard definition~\cite{goodfellow2014generative,mao2017least}:
\begin{equation}
\mathcal{L}_\text{adv} =  
\mathbb{E}\big[\log \big(\phi_\text{dis}(\mathbf{J}_\text{ef})\big)\big]\label{eq:loss_gan1}
+\mathbb{E}\big[\log \big(1-\phi_\text{dis}(\mathbf{J}_\text{refine})\big)\big].
\end{equation}
While our light-effects suppression network is designed to refine $\mathbf{J}_\text{init}$ by suppressing any remaining light effects, we also encourage it to output the same light-effects-free image when the input has no light-effects  $\mathbf{J}_\text{ef}$. We achieve this by using the following identity loss function~\cite{Zhu17}:
\begin{equation}
\mathcal{L}_\text{iden} =  
\mathbb{E} \big[\lVert\phi_\text{gen}(\mathbf{J}_\text{ef})-\mathbf{J}_\text{ef}\rVert_{1}\big].
\label{eq:loss_identity1}
\end{equation}

\noindent We multiply each loss function with its respective weight, we adjust $\lambda_\text{gray-feat}=1$,
$\lambda_{\text{atten}}=0.5$ with the same scale, and employ the weights of $\lambda_\text{adv}=1$ and $\lambda_\text{iden}=5$ from~\cite{Zhu17}. The HF layers use smoothing kernels ${K}$, with size given by $k= 2^i$, $i= 2,3,4,...$, the regularization $\epsilon= 0.04,0.08$.

\begin{table}[!t]
	\centering
	\renewcommand{\arraystretch}{1.2}
	\caption{User study evaluation on the real night data, our method obtained the highest mean and lowest standard deviation (the max score is 7), showing our method is realistic, light-effects (L.E.) suppressed, and has good visibility.}
	\resizebox{0.8\columnwidth}{!}{
		\begin{tabular}{l|cc|ccc|c|c}\hline 
			Three Aspects &EG~\cite{jiang2021enlightengan} &Afifi~\cite{afifi2021learning} &Yan~\cite{yan2020nighttime} &Zhang~\cite{zhang2017fast} &Li~\cite{li2015nighttime} &Sharma~\cite{sharma2021nighttime} & Ours\\\hline 
			1.Realism$\uparrow$           &$3.3\pm1.5$ &$5.5\pm1.3$  &$3.7\pm2.0$   &$3.5\pm1.6$ &$3.1\pm1.8$ &$2.8\pm1.5$ &$\bf 6.1\pm0.8$\\\hline
			2.L.E. Supp.$\uparrow$        &$1.7\pm0.8$ &$3.1\pm1.3$  &$4.6\pm1.4$   &$3.9\pm1.1$ &$5.2\pm1.2$ &$3.0\pm1.5$ &$\bf 6.6\pm0.7$\\\hline
			3.Visibility$\uparrow$        &$3.1\pm1.6$ &$4.2\pm1.5$  &$4.7\pm1.5$   &$3.7\pm1.1$ &$3.8\pm1.5$ &$3.0\pm1.4$ &$\bf 6.4\pm0.7$\\\hline
	\end{tabular}}
	\label{tb:tb_user}
\end{table}

\section{Experimental Results}
\label{sec:experiments}
\noindent \textbf{Light-Effects Suppression on Night Data}
The real night images used in our experiment are downloaded from the Internet and collected by ourselves. 
We use these images for our unpaired training since collecting the corresponding light-effects-free ground truth images is difficult. 

For the user study, we randomly selected 210 outputs (30 per method, seven methods) and presented them to the 12 participants in random order. 
We asked them to rank these methods from unrealistic (1) to realistic (7); light effects still present (1) to suppressed (7); poor visibility (1) to good visibility (7).
Table~\ref{tb:tb_user} shows the user study results.
Table~\ref{tb:table_quan} shows the quantitative results on the night data, where our method has the highest PSNR and SSIM scores.

\begin{figure}[t!]
	\captionsetup[subfloat]{labelformat=empty}
	\captionsetup[subfloat]{farskip=1pt}
	\centering
	\subfloat{\includegraphics[width = 0.195\columnwidth]{figures/night_data/y/yellow_input.jpg}}\hfill
	\subfloat{\includegraphics[width = 0.195\columnwidth]{figures/night_data/y/yellow_ours.jpg}}\hfill
	\subfloat{\includegraphics[width = 0.195\columnwidth]{figures/night_data/y/yellow_CVPR21hdr.jpg}}\hfill
	\subfloat{\includegraphics[width = 0.195\columnwidth]{figures/night_data/y/yellow_enlighten.jpg}}\hfill
	\subfloat{\includegraphics[width = 0.195\columnwidth]{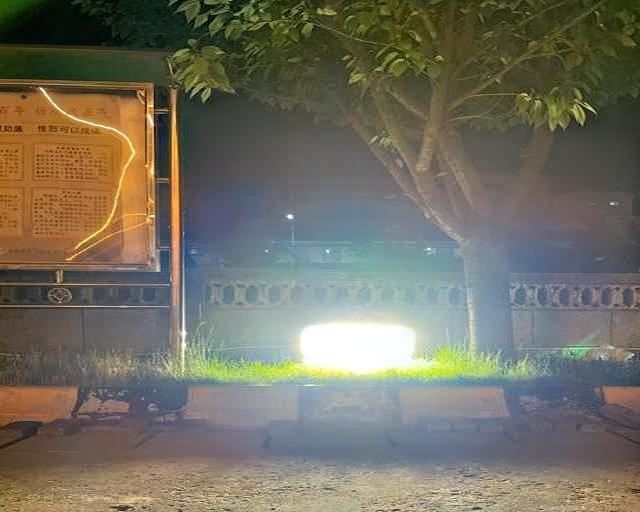}}\hfill
	\subfloat{\includegraphics[width = 0.195\columnwidth]{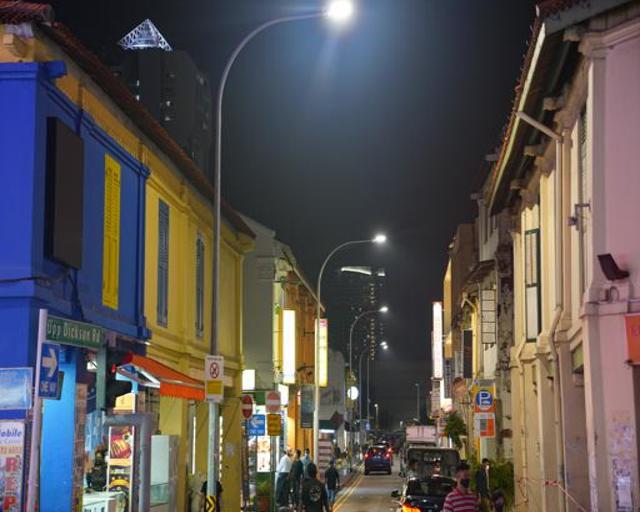}}\hfill
	\subfloat{\includegraphics[width = 0.195\columnwidth]{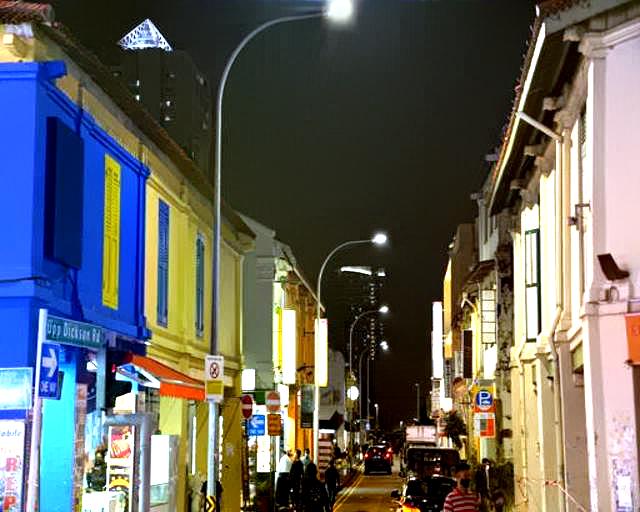}}\hfill
	\subfloat{\includegraphics[width = 0.195\columnwidth]{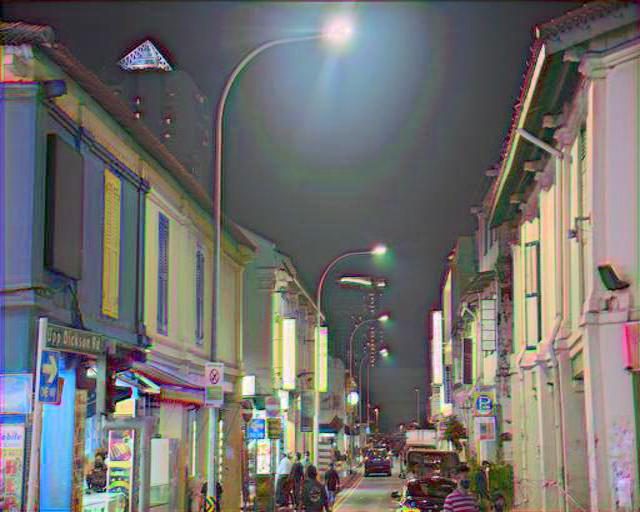}}\hfill
	\subfloat{\includegraphics[width = 0.195\columnwidth]{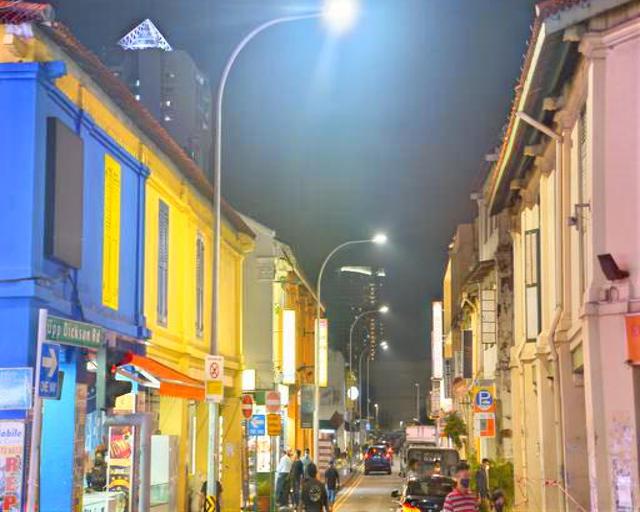}}\hfill
	\subfloat{\includegraphics[width = 0.195\columnwidth]{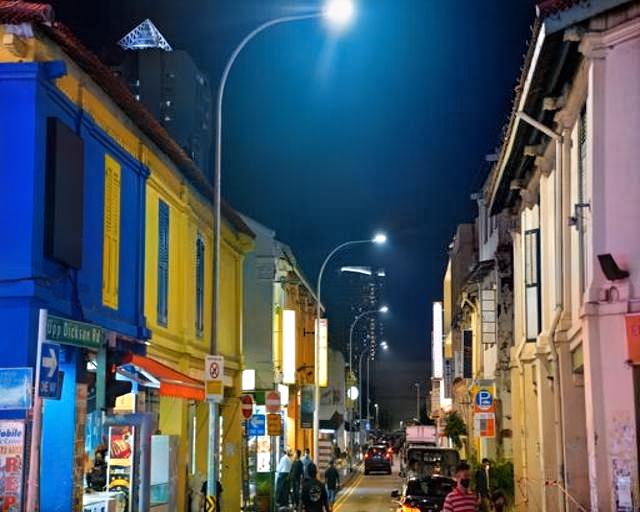}}\hfill
	\subfloat{\includegraphics[width = 0.195\columnwidth]{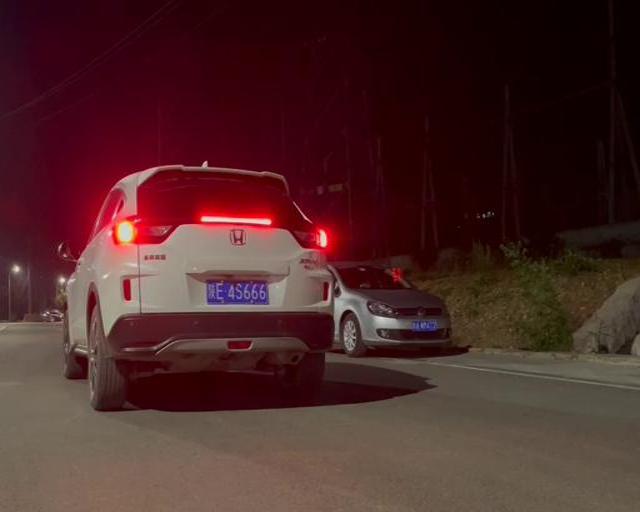}}\hfill
	\subfloat{\includegraphics[width = 0.195\columnwidth]{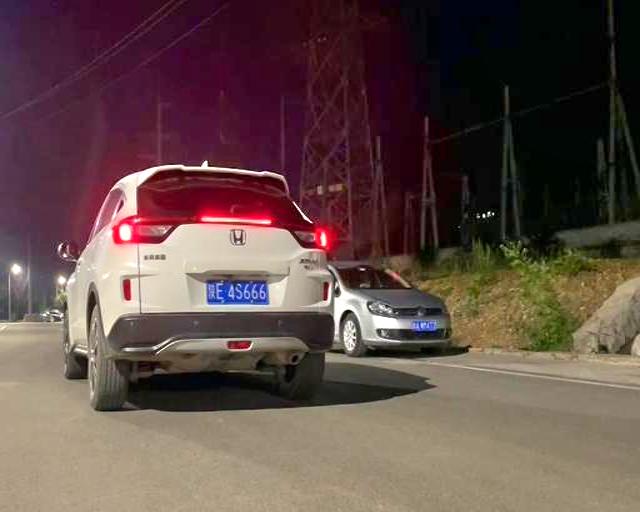}}\hfill
	\subfloat{\includegraphics[width = 0.195\columnwidth]{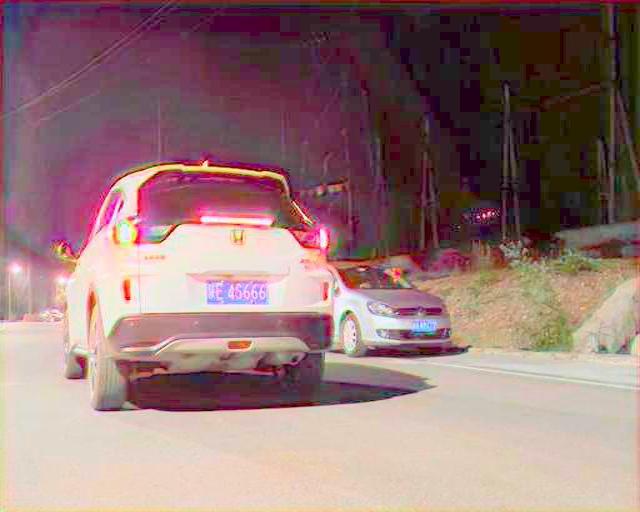}}\hfill
	\subfloat{\includegraphics[width = 0.195\columnwidth]{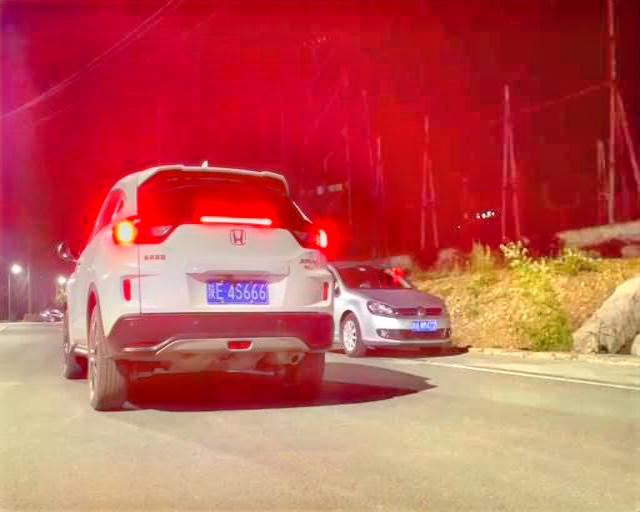}}\hfill
	\subfloat{\includegraphics[width = 0.195\columnwidth]{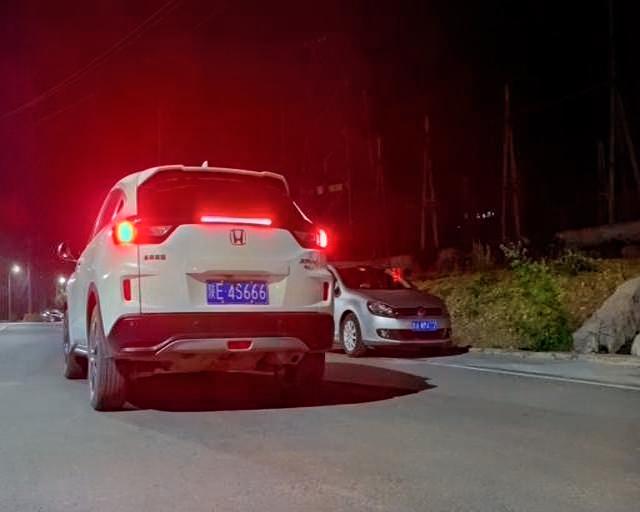}}\hfill	
	\subfloat[Input]{\includegraphics[width = 0.195\columnwidth]{figures/night_data/g/grass_input.jpg}}\hfill
	\subfloat[Ours]{\includegraphics[width = 0.195\columnwidth]{figures/night_data/g/grass_ours.jpg}}\hfill
	\subfloat[Sharma~\cite{sharma2021nighttime}]{\includegraphics[width = 0.195\columnwidth]{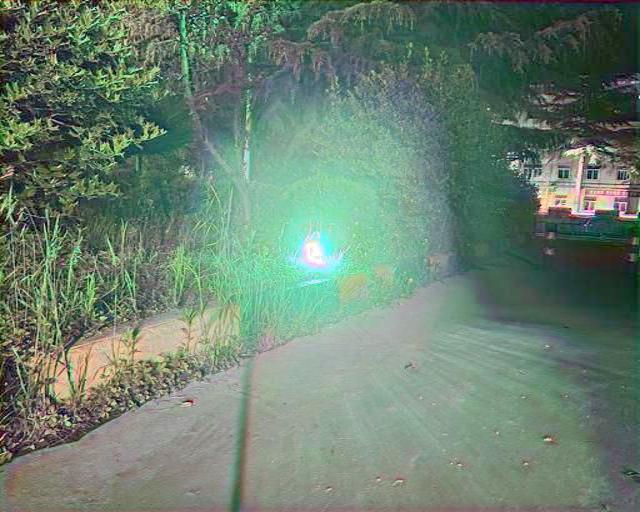}}\hfill
	\subfloat[EG~\cite{jiang2021enlightengan}]{\includegraphics[width = 0.195\columnwidth]{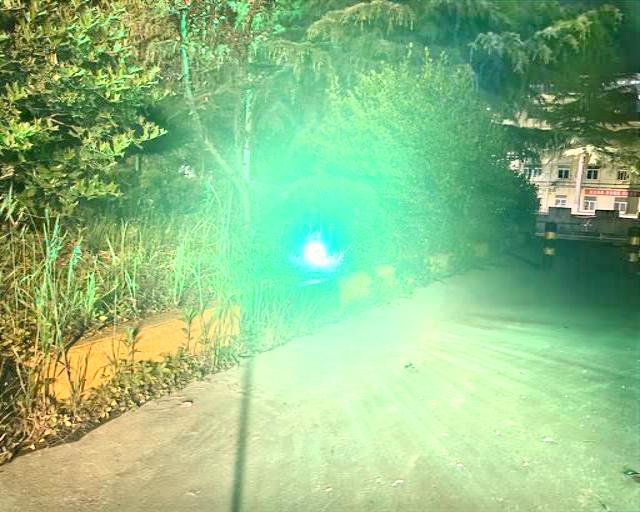}}\hfill
	\subfloat[Afifi~\cite{afifi2021learning}]{\includegraphics[width = 0.195\columnwidth]{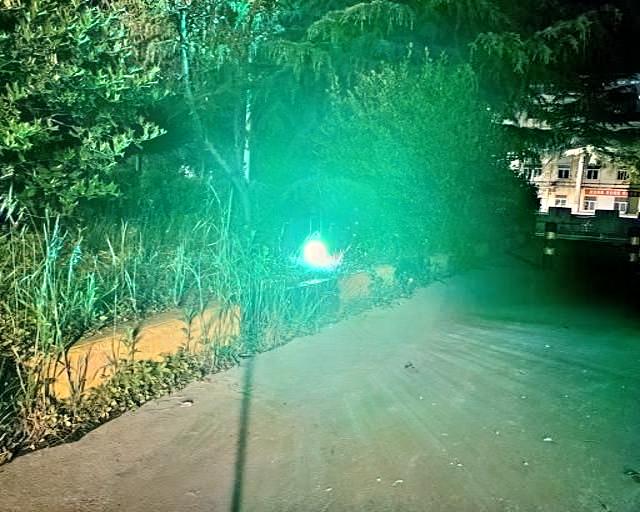}}\hfill
	\caption{Comparing light-effects suppression and dark regions enhancement results on the real night images.}\label{figure_ours_night_data}
\end{figure}

\begin{table}[!t]
	\centering
	\renewcommand{\arraystretch}{1.2}
	\caption{Quantitative light-effects suppression comparison on the night data. In the table, UL = unsupervised learning, SL = supervised learning, SSL = semi-supervised learning, ZSL = zero-shot learning, Opti = optimization method.}
	\resizebox{1\columnwidth}{!}{
		\begin{tabular}{c|c|cccc|ccc|c|c}\hline	
			Learning		&-	 &UL  &ZSL &SL 	&SL 	&SSL  		&Opti &Opti        &SSL &UL\\\hline
			Datasets &Metrics &EG~\cite{jiang2021enlightengan}  &ZD+~\cite{li2021learning} &RN~\cite{chen2018retinex} &Afifi~\cite{afifi2021learning} &Yan~\cite{yan2020nighttime} &Zhang~\cite{zhang2017fast} &Li~\cite{li2015nighttime} &Sharma~\cite{sharma2021nighttime} &Ours \\\hline
			\multirow{2}{*}{GTA5~\cite{yan2020nighttime}}	  &PSNR$\uparrow$ &10.94 &21.13 &7.79 &15.47  &26.99 &20.92 &21.02 &8.14 &\bf29.79\\			                  
			&SSIM$\uparrow$ &0.31  &0.68 &0.23 &0.53  &0.85  &0.65  &0.64 &0.29 &\bf 0.88 \\\hline
			\multirow{2}{*}{Syn-light-effects~\cite{metari2007new}} 	  &PSNR$\uparrow$ &7.38  &7.84  &6.39 &11.31  &14.88 &16.30 &14.66  &14.00 &\bf16.95\\
			&SSIM$\uparrow$ &0.17  &0.20  &0.16 &0.35  &0.23  &0.38  &0.37  &0.37 &\bf0.39\\\bottomrule
		\end{tabular}
	}
	\label{tb:table_quan}
\end{table}

\begin{figure}[t!]
	\captionsetup[subfloat]{labelformat=empty}
	\captionsetup[subfloat]{farskip=1pt}
	\centering	
	\subfloat[Input]{\includegraphics[width = 0.195\columnwidth]{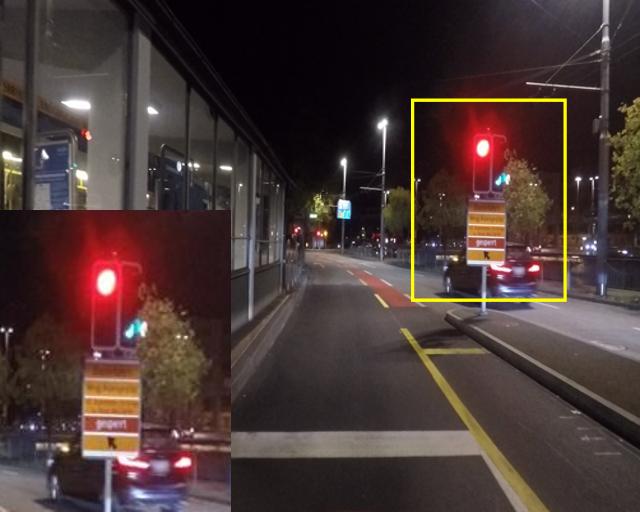}}\hfill
	\subfloat[Ours]{\includegraphics[width = 0.195\columnwidth]{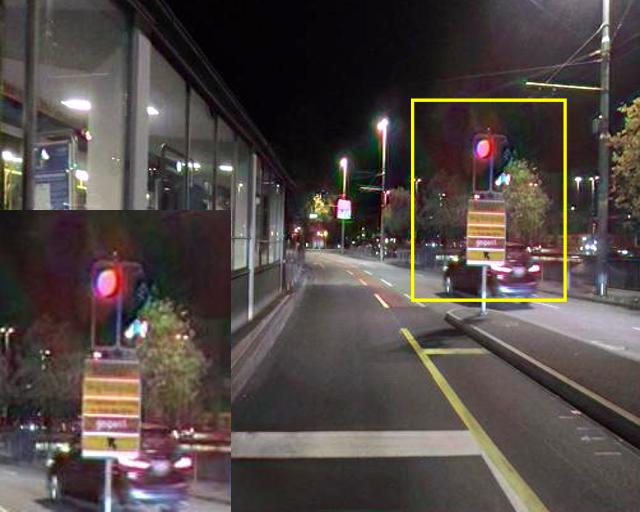}}\hfill
	\subfloat[Sharma~\cite{sharma2021nighttime}]{\includegraphics[width = 0.195\columnwidth]{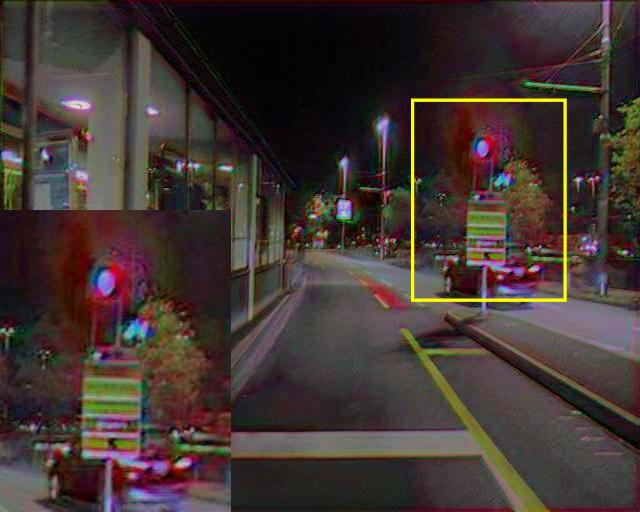}}\hfill
	\subfloat[EG~\cite{jiang2021enlightengan}]{\includegraphics[width = 0.195\columnwidth]{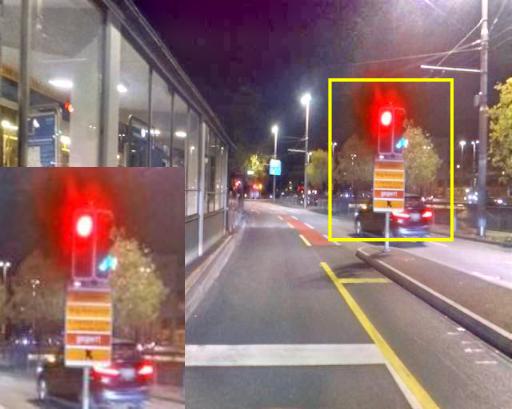}}\hfill
	\subfloat[Afifi~\cite{afifi2021learning}]{\includegraphics[width = 0.195\columnwidth]{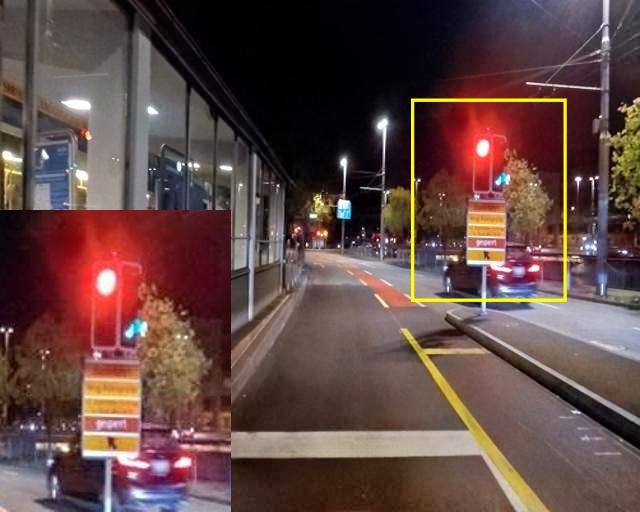}}\hfill
	\caption{Comparing light-effects suppression and dark regions enhancement results on the real night image from Dark Zurich~\cite{sakaridis2019guided} dataset.}\label{figure_public_night_data}
\end{figure}

Fig.~\ref{figure_ours_night_data} shows the qualitative results on real night images, which demonstrate the superiority of our results compared to the baseline methods.
Fig.~\ref{figure_public_night_data} shows the evaluation on the Dark Zurich~\cite{sakaridis2019guided} dataset.
As can be observed, the light-effects suppression baseline~\cite{sharma2021nighttime} suffers from hallucination/artefacts and cannot handle white light effects.
In the supplementary material, we show the results of night dehazing baselines~\cite{yan2020nighttime,zhang2017fast,li2015nighttime}, which are too dark since they are not designed to enhance dark regions;
while low-light image enhancement baselines~\cite{jiang2021enlightengan,afifi2021learning,li2021learning,chen2018retinex} wrongly intensify light effects, and thus degrade the visibility of the images.

\begin{table}[t!]
	\centering
	\renewcommand{\arraystretch}{1.2}
	\caption{Quantitative comparisons on the LOL-test dataset~\cite{chen2018retinex}.}
	\resizebox{0.6\columnwidth}{!}{
		\begin{tabular}{c|l|cccc}\hline 
			\multirow{2}{*}{ Learning } & \multirow{2}{*}{ Method } & \multicolumn{4}{c}{ LOL-test } \\
			\cline{3-6} 
			& & MSE$\left(\times 10^{3}\right)$$\downarrow$ & PSNR$\uparrow$ & SSIM$\uparrow$ & LPIPS$\downarrow$\\\hline 
	Opti	& LIME~\cite{guo2016lime}     				  &	-    &16.760 &0.560 &0.350 \\\hline	
			& RetinexNet~\cite{chen2018retinex}           &1.651 &16.774 &0.462 &0.474 \\
			& KinD++~\cite{zhang2021beyond}               &1.298 &17.752 &0.760 &\bf{0.198} \\
	SL		& Afifi~\cite{afifi2021learning}              &4.520    &15.300 &0.560	&0.392     \\
			& RUAS~\cite{liu2021retinex} 				  &3.920	&18.230 &0.720 &0.350\\\hline			
		ZSL & ZeroDCE~\cite{guo2020zero}                  &3.282 &14.861 &0.589 &0.335 \\\hline			
		SSL & DRBN~\cite{yang2020fidelity}                &2.359 &15.125 &0.472 &0.316 \\
		UL  & EnlightenGAN~\cite{jiang2021enlightengan}   &1.998 &17.483 &0.677 &0.322 \\\hline
			
		SSL & Sharma~\cite{sharma2021nighttime}           &3.350   &16.880 &0.670 &0.315    \\\hline
		UL	& Ours		                                  &\bf{1.070} &\bf{21.521} &\bf{0.763} &0.235 \\\hline 
		\end{tabular}
	}
	\label{tb:lol}
\end{table}

\begin{table*}[t!]
	\centering
	\caption{Quantitative comparisons on the \textit{LOL-Real} dataset~\cite{yang2021sparse}.}
	\resizebox{1\columnwidth}{!}{
		\begin{tabular}{c|cccccccccccccc}\hline		
			Learning & NA    & Opti   & Opti   & Opti &ZSL &ZSL &ZSL &ZSL &SL\\ 
			Method   &Input &JED~\cite{ren2018joint} &RRM~\cite{li2018structure} &SRIE~\cite{fu2016weighted} &RDIP~\cite{zhao2021retinexdip}&MIRNet~\cite{Zamir2020MIRNet}  &RRDNet~\cite{zhu2020zero} &ZD~\cite{guo2020zero} &RUAS~\cite{liu2021retinex}\\
			
			PSNR$\uparrow$     &9.72 &17.33 &17.34 &17.34 &11.43 &12.67  &14.85 &20.54 &15.33\\
			SSIM$\uparrow$     &0.18  &0.66  &0.68  &0.68  &0.36  &0.41   &0.56  &0.78  &0.52\\\hline
			
			Learning &SL   & SL      & SL     &  SL   &SL  & SSL 	     &UL &SSL &UL\\
			Method   &LLNet~\cite{lore2017llnet} &RN~\cite{chen2018retinex} &DUPE~\cite{Wang_2019_CVPR}  &SICE~\cite{Cai2018deep} &Afifi~\cite{afifi2021learning} &DRBN~\cite{yang2021band}   &EG~\cite{jiang2021enlightengan} &Sharma~\cite{sharma2021nighttime} &Ours\\
			PSNR$\uparrow$ &17.56    & 15.47 & 13.27  &19.40 &16.38 &19.66  &18.23 &18.34 &\textbf{25.51}\\
			SSIM$\uparrow$ &0.54     & 0.56  & 0.45   &0.69  &0.53 &0.76 &0.61  &0.64 &\bf{0.80}\\
			\bottomrule 
	\end{tabular}}
	\label{tb:lol_real}
\end{table*}

\subsubsection{Low-Light Enhancement} 
Besides night light-effects suppression, our method can boost the brightness of low light images with no light effects, by simply setting the light-effects layer to $\mathbf{G}_\text{0}$, which has no light-effects.
For a fair comparison, we compare low-light boosting with image enhancement methods without considering the presence of light effects.

\begin{figure*}[t]
	\captionsetup[subfigure]{font=small, labelformat=empty}
	\captionsetup[subfloat]{farskip=1pt}
	\centering
	\subfloat{\includegraphics[width = 0.195\columnwidth]{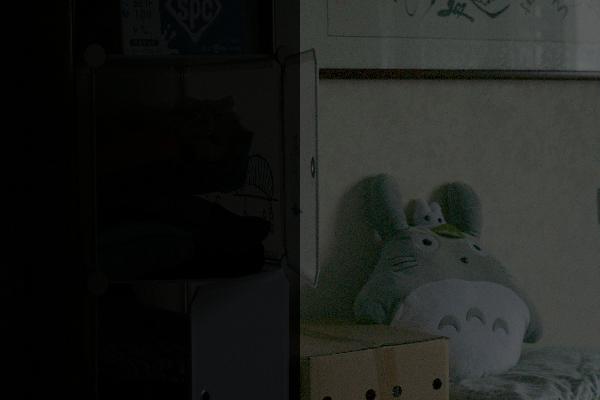}}\hfill
	\subfloat{\includegraphics[width = 0.195\columnwidth]{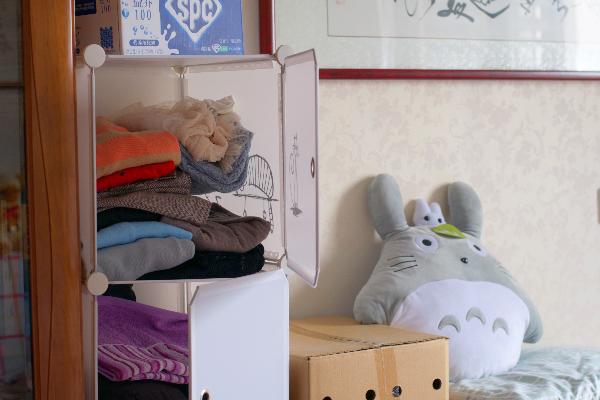}}\hfill
	\subfloat{\includegraphics[width = 0.195\columnwidth]{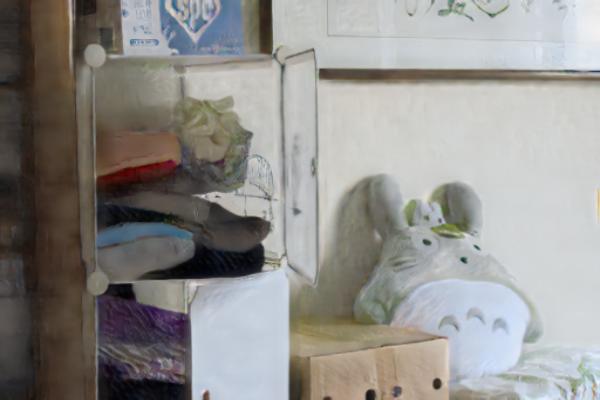}}\hfill
	\subfloat{\includegraphics[width = 0.195\columnwidth]{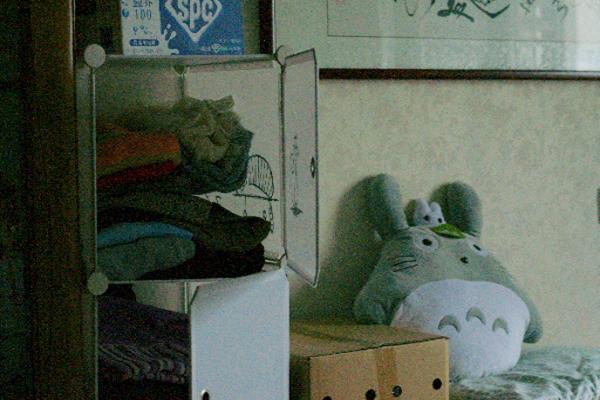}}\hfill
	\subfloat{\includegraphics[width = 0.195\columnwidth]{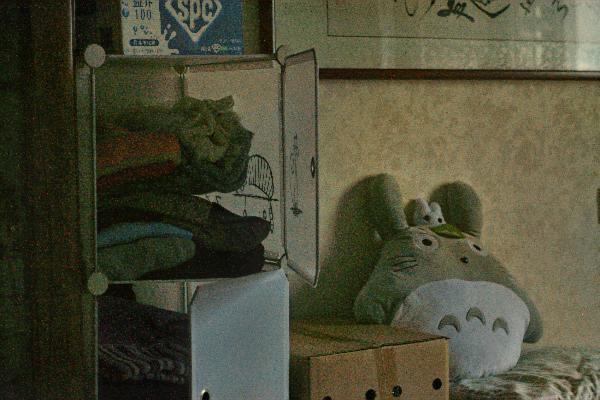}}\hfill	
	\subfloat[Input]{\includegraphics[width = 0.195\columnwidth]{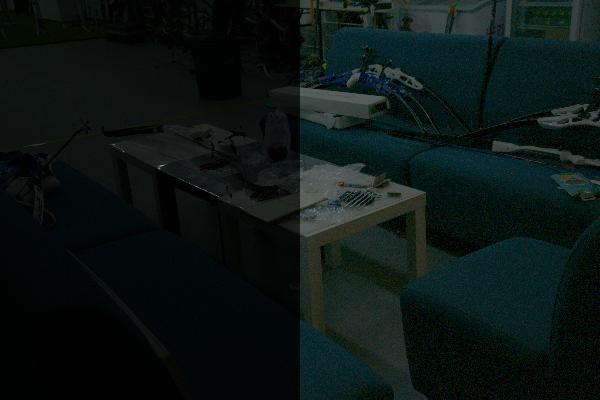}}\hfill
	\subfloat[Ground Truth]{\includegraphics[width = 0.195\columnwidth]{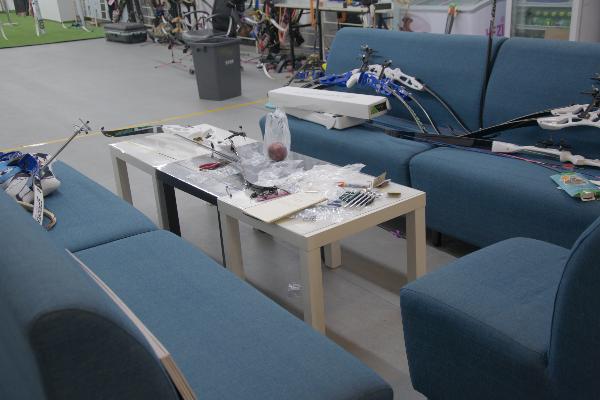}}\hfill
	\subfloat[Ours]{\includegraphics[width = 0.195\columnwidth]{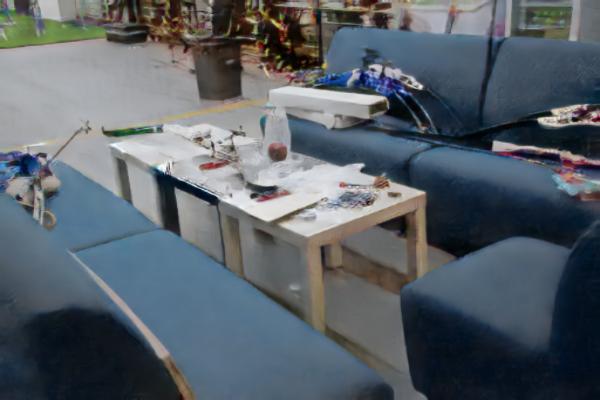}}\hfill
	\subfloat[Sharma~\cite{sharma2021nighttime}]{\includegraphics[width = 0.195\columnwidth]{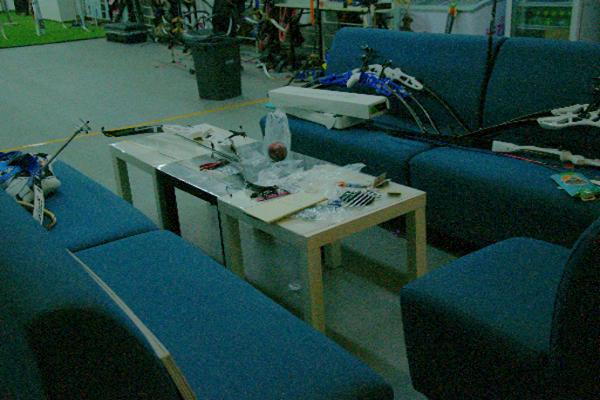}}\hfill
	\subfloat[EG~\cite{jiang2021enlightengan}]{\includegraphics[width = 0.195\columnwidth]{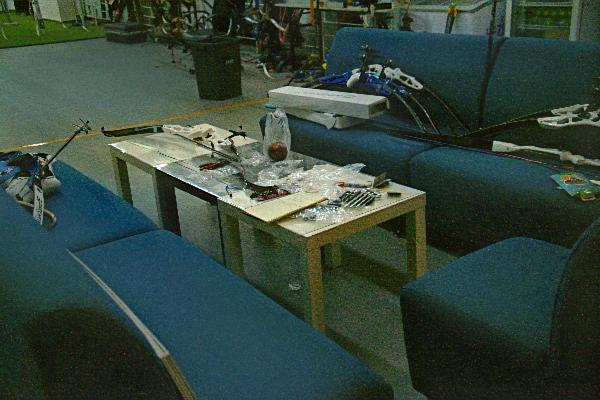}}\hfill
	\caption{Low-light enhancement results on the LOL-test~\cite{chen2018retinex}, \textit{LOL-Real}~\cite{yang2021sparse} datasets.}
	\label{figure_low_light}
\end{figure*}

We adopt the LOL dataset~\cite{chen2018retinex}\footnote{The LOL dataset link: \url{https://daooshee.github.io/BMVC2018website/}}, 485 training and 15 testing images, respectively.
Table~\ref{tb:lol} shows quantitative results, where our method achieves better performance compared with the baseline methods in terms of PSNR, SSIM, Mean Square Error (MSE) and Learned Perceptual Image Patch Similarity (LPIPS)~\cite{zhang2018unreasonable}.
We evaluate on \textit{LOL-Real}~\cite{yang2021sparse}\footnote{\textit{LOL-Real} dataset link: \url{https://github.com/flyywh/CVPR-2020-Semi-Low-Light/}}, 100 testing images with more diversified scenes. 
We train our method on the LOL dataset and test on the \textit{LOL-Real} test-split. 
The results are shown in Table~\ref{tb:lol_real} and Fig.~\ref{figure_low_light}, showing the generality of our method.
Our method achieves better performance compared with the baseline methods in terms of PSNR, SSIM.

\noindent \textbf{Baselines}
As shown in Table~\ref{tb:compare}, there is only one algorithm, i.e., Sharma and Tan~\cite{sharma2021nighttime} that suppresses night light effects and boosts the dark regions simultaneously. Yet, the method cannot handle white light effects and suffers from hallucination/artefacts.
Night dehazing methods can suppress glow, but are suboptimal to enhancing low-light regions. Low-light image enhancement methods are not designed to suppress night light effects and enhance low light regions simultaneously.

Nevertheless for comprehensive comparisons, besides comparing with~\cite{sharma2021nighttime}, we also compare our method with the state-of-the-art single-image low-light image enhancement methods: EnlightenGAN~\cite{jiang2021enlightengan}, Afifi et al.~\cite{afifi2021learning}, etc. and the night dehazing methods: Yan et al.~\cite{yan2020nighttime}, Zhang et al.~\cite{zhang2017fast}, Li et al.~\cite{li2015nighttime}, etc.
The codes of all the baseline methods are obtained from the authors.
More baseline results are provided in the supplementary material.

\begin{table}[!t]
	\centering
	\renewcommand{\arraystretch}{1.2}
	\caption{Summary of comparisons between our method and existing night image enhancement methods. Our method can suppress light effects (including white light effects), preserve light source (L.S.) details, and boost dark regions simultaneously.}
	\resizebox{0.82\columnwidth}{!}{
		\begin{tabular}{c|l|c|c|c|c}\hline
			Learning &Methods &\multicolumn{3}{c|}{Light Effects (L.E.) Suppression} &\multicolumn{1}{c}{Dark Regions Boosting}\\\cline{3-5} 
			&        & Normal L.E. & White L.E. &Details in L.S. &\\\hline
			{\multirow{1}{*}{UL}}   &Ours	   &\checkmark  &\checkmark &\checkmark &\checkmark\\\hline
			{\multirow{1}{*}{SSL}} &Sharma and Tan~\cite{sharma2021nighttime}   & \checkmark  & $\times$ & $\times$ &\checkmark\\\hline
			{\multirow{1}{*}{Opti;SSL}} &Night Dehazing~\cite{li2015nighttime,zhang2017fast,yan2020nighttime} &\checkmark  &\checkmark &$\times$ &$\times$\\\hline 
			{\multirow{1}{*}{SL;ZSL;UL}} & Low-light Enhancement~\cite{afifi2021learning,li2021learning,jiang2021enlightengan}  &\multicolumn{3}{c|}{$\times$}     &\checkmark\\\hline                              										 
	\end{tabular}}
	\label{tb:compare}
\end{table}

\begin{figure}[t!]
	\captionsetup[subfloat]{farskip=1pt}
	\setcounter{subfigure}{0}
	\subfloat[]{\includegraphics[width = 0.163\columnwidth]{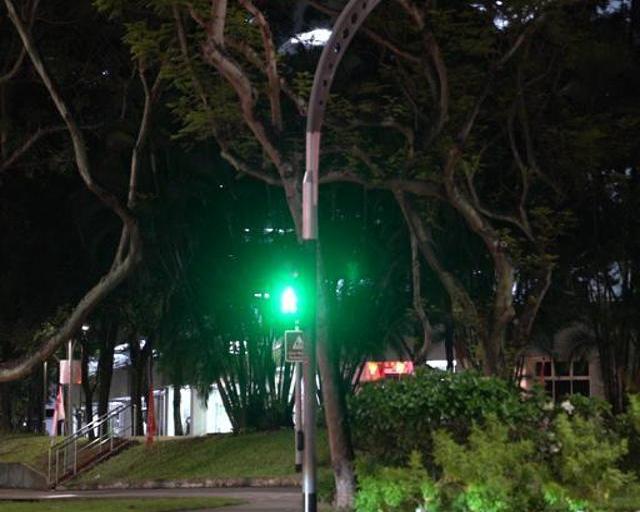}}\hfill
	\subfloat[]{\includegraphics[width = 0.163\columnwidth]{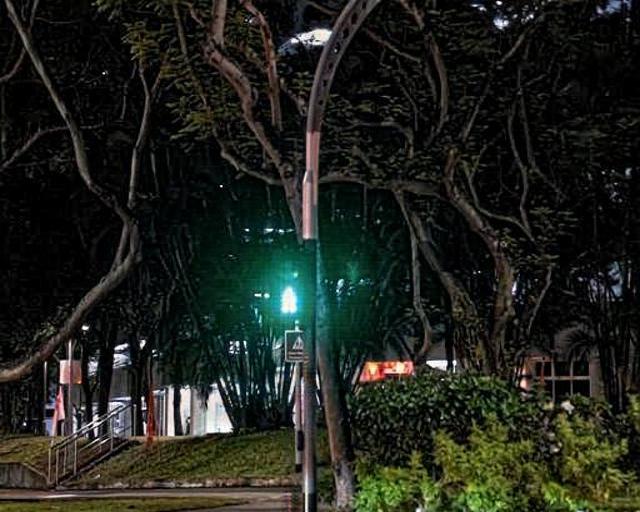}}\hfill
	\subfloat[\label{fig_SB}]{\includegraphics[width = 0.163\columnwidth]{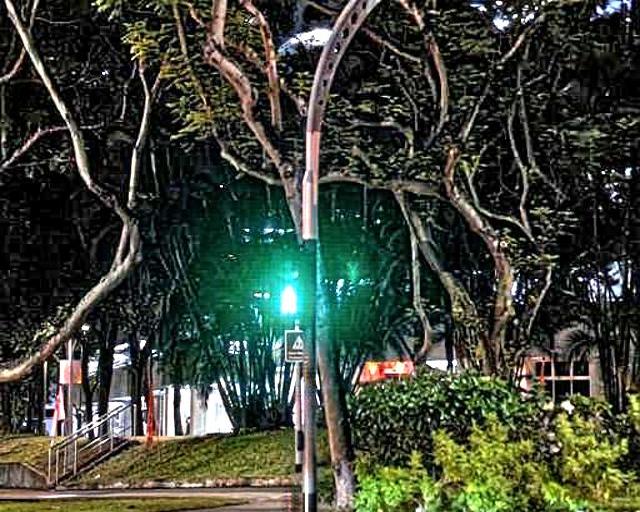}}\hfill	
	\subfloat[]{\includegraphics[width=0.163\columnwidth]{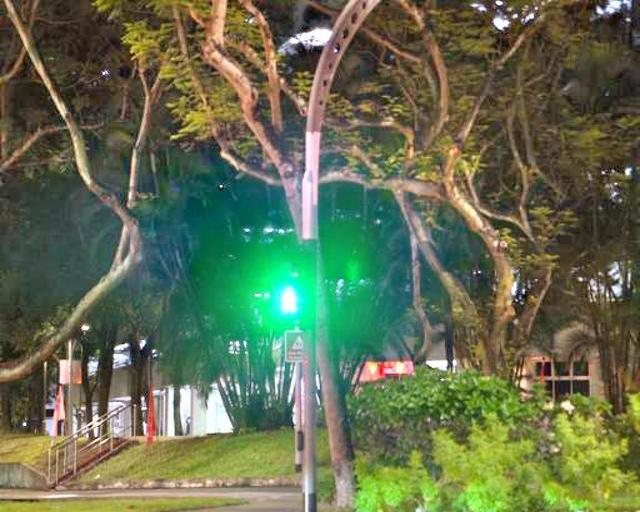}}\hfill
	\subfloat[\label{fig_BS}]{\includegraphics[width=0.163\columnwidth]{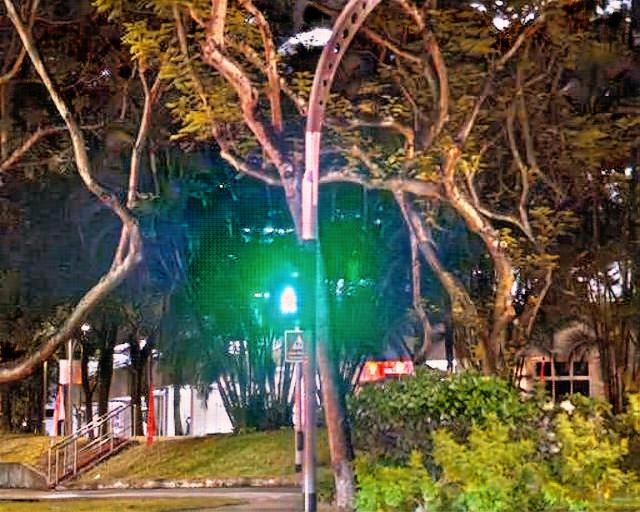}}\hfill
	\subfloat[]{\includegraphics[width = 0.163\columnwidth]{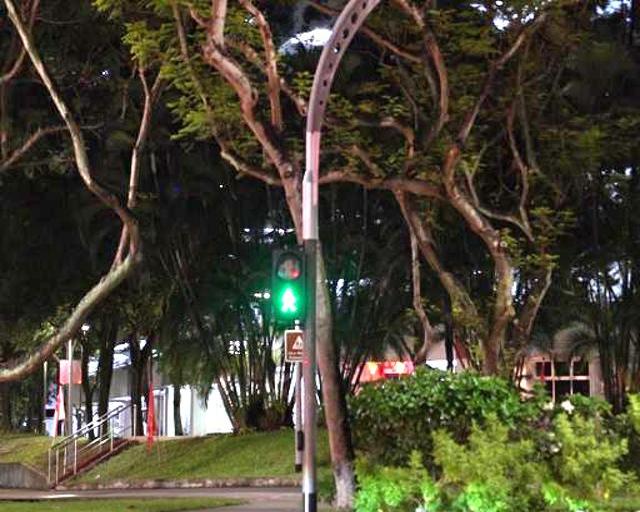}}\hfill
	\caption{Experiments on the effectiveness of joint light-effects suppression and  dark regions boosting:  (a) input, (b) light-effects suppression, (c) light-effects suppression followed by boosting without joint training, (d) boosting, (e) boosting followed by light-effects suppression without joint training, (f) our joint training light-effects suppression and boosting.}
	\label{figure_joint}
\end{figure}

\noindent \textbf{Joint Light-Effects Suppression and Dark Region Boosting}
As shown in Fig.~\ref{figure_joint}, jointly suppressing light-effects and then boosting dark regions are more effective than any other possibilities (namely, (b) the light-effects suppression alone, (c) light-effects suppression followed by boosting without jointly training them, (d) boosting alone, and (e) boosting followed by light effects suppression without joint training).
If we suppress light effects first, then boost the intensity without the joint training, as shown in Fig.~\ref{fig_SB}, artefacts and remaining light effects are also enhanced.
If we boost the intensity first, then suppress light effects without joint training, as shown in Fig.~\ref{fig_BS}, light effects cannot be effectively suppressed since the amplified light effects cause information and detail loss.

\subsubsection{Ablation Studies}
\label{sec_ablation}
\begin{figure}[t!]
	\captionsetup[subfloat]{labelformat=empty}
	\captionsetup[subfloat]{farskip=1pt}
	\centering		
	\subfloat[Input]{\includegraphics[width=0.245\textwidth,height=1.8cm]{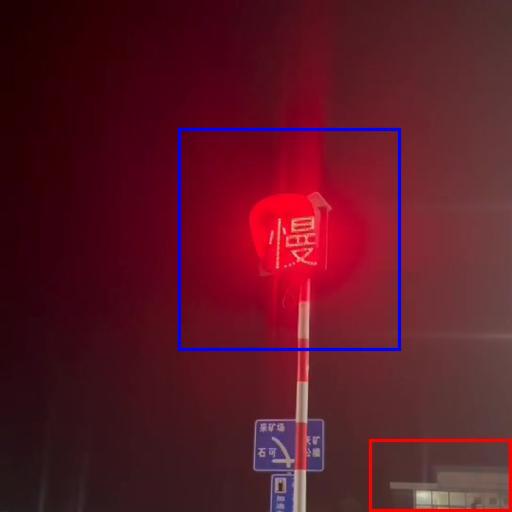}}\hfill		
	\subfloat[w/o Trans.]{\includegraphics[width=0.245\textwidth,height=1.8cm]{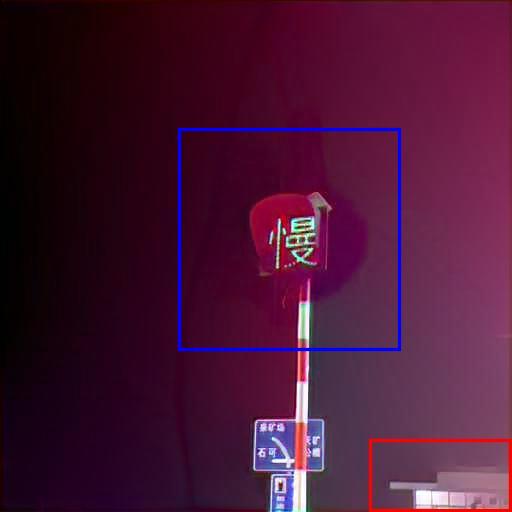}}\hfill
	\subfloat[w/o Decomp.]{\includegraphics[width=0.245\textwidth,height=1.8cm]{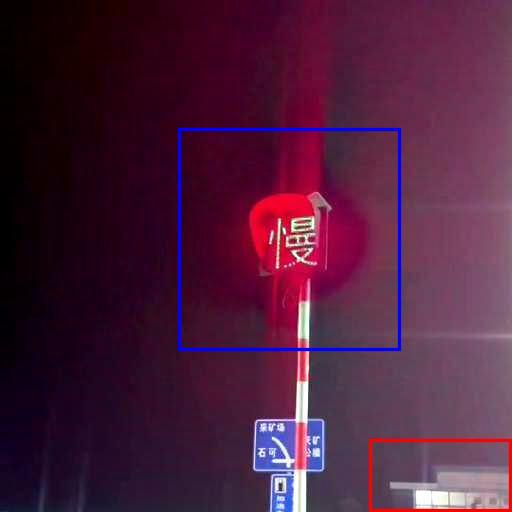}}\hfill
	\subfloat[Output]{\includegraphics[width=0.245\textwidth,height=1.8cm]{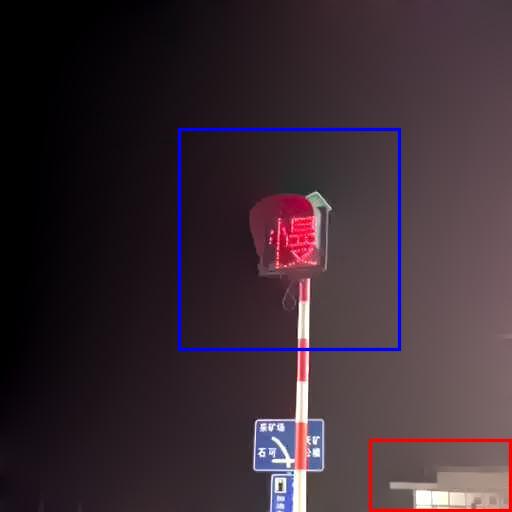}}\hfill
	\caption{Ablation studies on the framework. `w/o Trans' denotes our method without light-effects suppression. `w/o Decomp.' denotes our method without layer decomposition. We can observe that our framework is important for night image enhancement.}
	\label{figure_ablation}
\end{figure}

\begin{figure}[t]
	\captionsetup[subfloat]{labelformat=empty}
	\captionsetup[subfloat]{farskip=2pt}	
	\subfloat[Input]{\includegraphics[width=0.245\columnwidth,height=1.8cm]{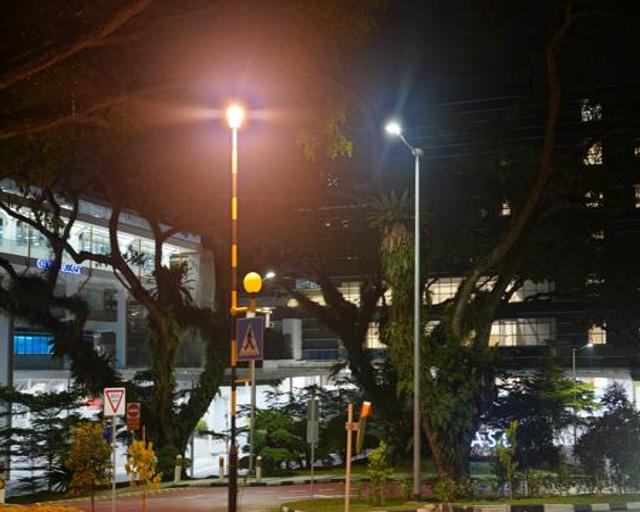}}\hfill
	\subfloat[Light-Effects Layer]{\includegraphics[width=0.245\columnwidth,height=1.8cm]{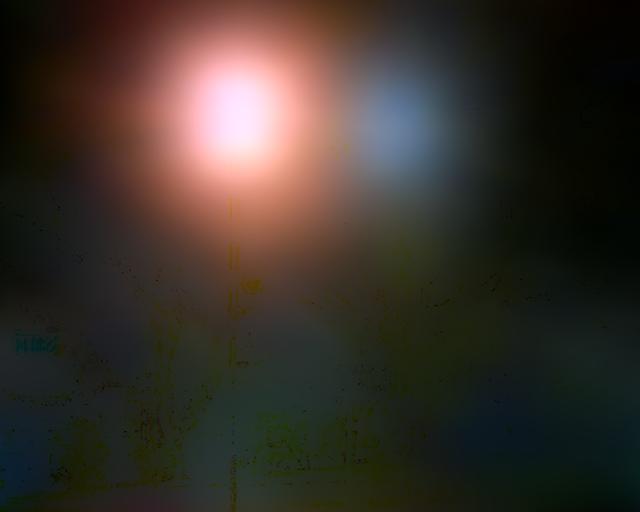}}\hfill
	\subfloat[w/o $\mathbf{G}$ guidance]{\includegraphics[width=0.245\columnwidth,height=1.8cm]{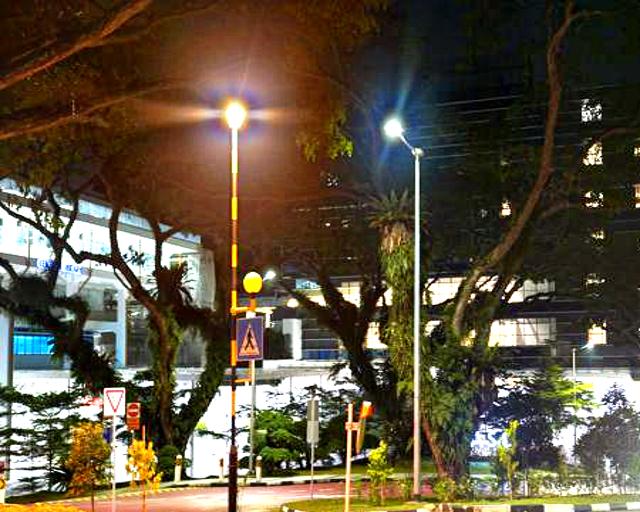}}\hfill	
	\subfloat[w/ $\mathbf{G}$ guidance]{\includegraphics[width=0.245\columnwidth,height=1.8cm]{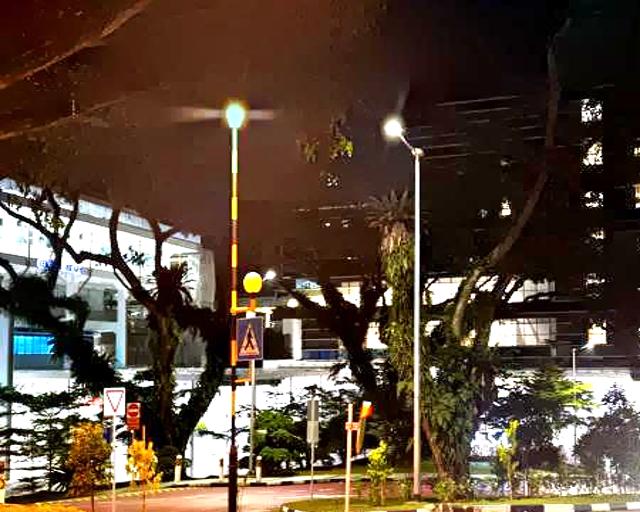}}\hfill 
	\caption{Ablation studies on light-effects layer guidance, with light-effects layer $\mathbf{G}$ guidance, our light-effects suppression network can focus on light-effects regions, separate light effects more properly.}
	\label{fig_ab_layer}
\end{figure}

\begin{figure}[t!]
	\captionsetup[subfloat]{labelformat=empty}
	\captionsetup[subfloat]{farskip=1pt}
	\subfloat[Input]{\includegraphics[width = 0.163\columnwidth]{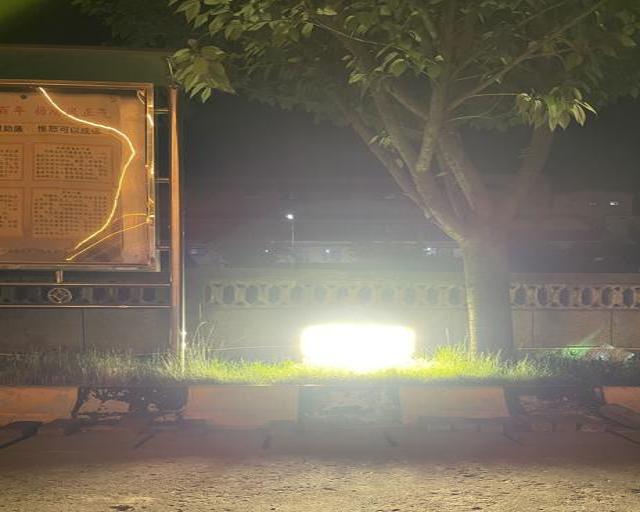}}\hfill
	\subfloat[w/o $\mathcal{L}_\text{gray-feat}$]{\includegraphics[width = 0.163\columnwidth]{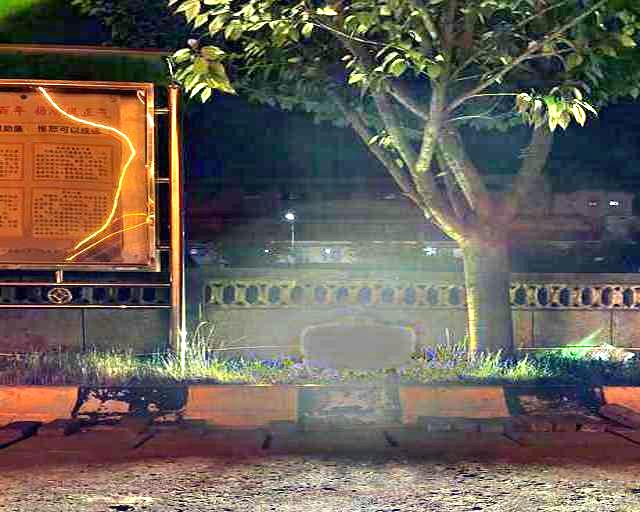}}\hfill	
	\subfloat[w/ $\mathcal{L}_\text{gray-feat}$]{\includegraphics[width=0.163\columnwidth]{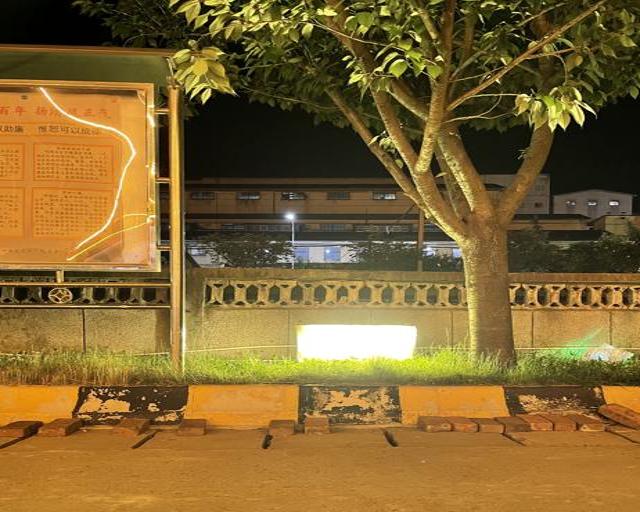}}\hfill
	\subfloat[Input]{\includegraphics[width = 0.163\columnwidth]{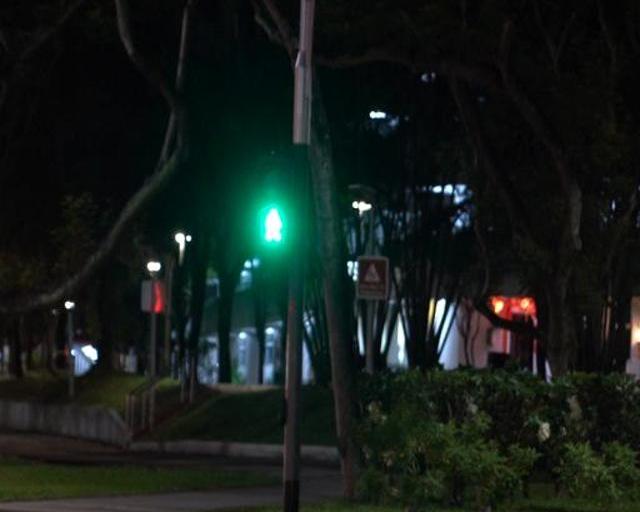}}\hfill
	\subfloat[w/o $\mathcal{L}_\text{gray-feat}$]{\includegraphics[width = 0.163\columnwidth]{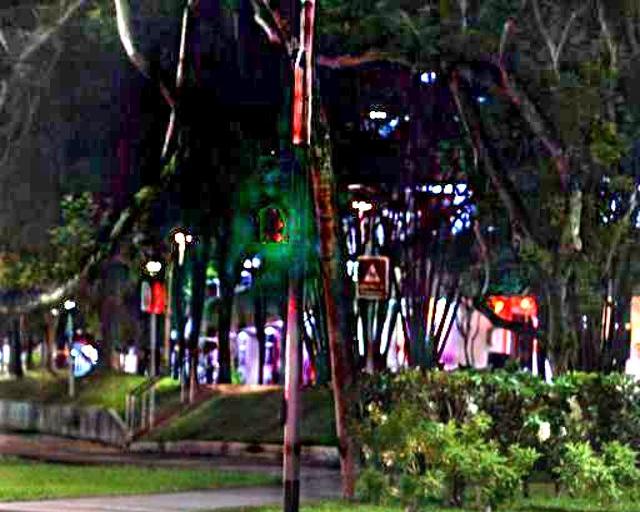}}\hfill
	\subfloat[w/ $\mathcal{L}_\text{gray-feat}$]{\includegraphics[width = 0.163\columnwidth]{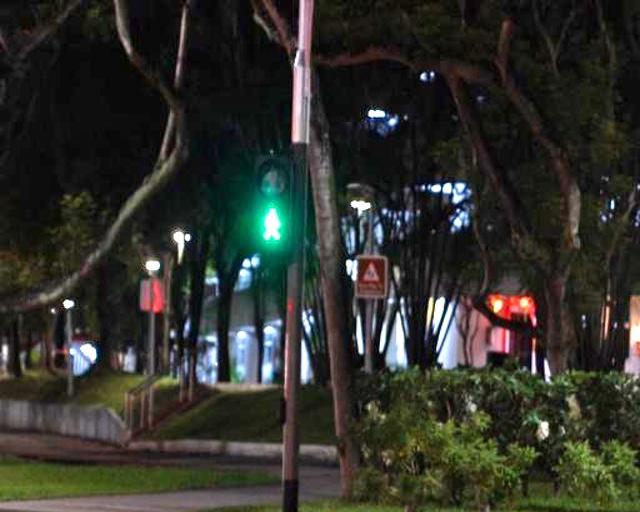}}\hfill\\
	\caption{Ablation studies on the structure and HF-features consistency losses $\mathcal{L}_\text{gray-feat}$, with $\mathcal{L}_\text{gray-feat}$, our method suppresses artefacts, and preserves details.}
\label{fig_ab_gray}
\end{figure}

Fig.~\ref{figure_ablation}, Fig.~\ref{fig_ab_layer} and Fig.~\ref{fig_ab_gray} show the effectiveness of our framework, light-effects layer guidance and structure and HF-features consistency losses used in our method, which clearly show that all the components are important for better performance.

\noindent \textbf{Decomposition + Suppression}
To show the effectiveness of our model-based unsupervised decomposition, we train our network without the decomposition module.
We directly input the night images to the light-effects suppression network, thus there is no light-effects layer guidance and initial background results.
Similarly, to show the effectiveness of our unsupervised light-effects suppression, we assume the initial background image $\mathbf{J_\text{init}}$ generated by the decomposition part is the final result without any refinement. 
Our final results are more effective in suppressing light effects and more natural in recovering the background. 

\noindent \textbf{Light-Effects Layer Guidance}
We compare the results by our method with and without light-effects layer guidance.
Instead of input $\left(\mathbf{G}, \mathbf{J}_\text{init}\right)$, we input $\left(\mathbf{G}_\text{0}, \mathbf{J}_\text{init}\right)$ to the light-effects suppression network.
That means there is no light-effects layer $\mathbf{G}$, we concatenate the initially estimated
background scene with all zero maps $\mathbf{G}_\text{0}$.
Figs.~\ref{figure_model}-\ref{figure_attention} show the results of the light-effects layer.
Fig.~\ref{fig_ab_layer} shows with light-effects layer guidance, our method can distinguish light-effects regions from background regions, focus on light-effects regions and properly suppress light effects (including white and multi-color light effects). 

\noindent \textbf{Structure and HF-features Consistency}
Structure and HF-features consistency losses can suppress artefacts and restore missing details.  
Fig.~\ref{fig_ab_gray} compares the results by our method with and without this loss.

\section{Conclusion}
\label{sec:conclusion}
In this paper, we have proposed a method to suppress light effects, and at the same time, boost the intensity of dark regions, from a single night image.
To achieve our goal, we cast the problem of light-effects suppression as an unsupervised decomposition problem.
We proposed an integrated network consisting of layer decomposition and light-effects suppression networks.
Our experiments show that our method outperforms the state-of-the-art visibility enhancement and light effects suppression methods. 

\section*{Acknowledgment}
This research/project is supported by the National Research Foundation, Singapore under its AI Singapore Programme (AISG Award No: AISG2-PhD/2022-01-037[T]), and partially supported by MOE2019-T2-1-130.
Wenhan Yang's research is supported by Wallenberg-NTU Presidential Postdoctoral Fellowship.	
Robby T. Tan's work is supported by MOE2019-T2-1-130.

\clearpage
\bibliographystyle{splncs04}
\bibliography{egbib}
\end{document}